\UseRawInputEncoding
\documentclass[11pt]{article}

\usepackage[preprint]{acl}

\usepackage{times}
\usepackage{latexsym}

\usepackage[T1]{fontenc}

\usepackage[utf8]{inputenc}

\usepackage{microtype}

\usepackage{inconsolata}
\usepackage{graphicx}
\usepackage{hyperref}
\usepackage{url}
\usepackage{xspace}
\usepackage{pifont}
\usepackage{graphicx}
\usepackage{booktabs}
\usepackage{wrapfig}
\usepackage{multirow}
\usepackage{multicol}
\usepackage{caption}
\usepackage{xcolor}
\usepackage{booktabs}
\usepackage{threeparttable}
\usepackage{colortbl}
\usepackage{makecell}
\usepackage[most,theorems]{tcolorbox}
\usepackage{tablefootnote}

\title{\dataname: Ambiguous Multi-hop Questions for Benchmarking LLM Answer Completeness}

\author{Jiabao Ji~\thanks{Work done when Jiabao was interning at Apple.}\\ UCSB\\ jiabaoji@ucsb.edu
    \And  Min Li  \\ Apple \\ min\_li6@apple.com
    \And Priyanshu Kumar \\ Apple \\ priyanshu\_kumar@apple.com
    \AND Shiyu Chang \\ UCSB \\ chang87@ucsb.edu
    \And Saloni Potdar \\ Apple \\ s\_potdar@apple.com
}

\newcommand{\dataname}{\textsc{DeepAmbigQA}\xspace}
\newcommand{\datagen}{\textsc{DeepAmbigQAGen}\xspace}

\begin{document}

\makeatletter
\lst@InstallKeywords k{attributes}{attributestyle}\slshape{attributestyle}{}ld
\makeatother

\definecolor{pythonblue}{rgb}{0.16,0.12,0.93}
\definecolor{cppgreen}{rgb}{0.16,0.42,0.16}
\definecolor{promptinsert}{HTML}{bfefff}
\definecolor{compcolor}{HTML}{90EE90}
\definecolor{codehlcolor}{HTML}{ffec8b}
\definecolor{codehlcolor2}{HTML}{ffbbff}
\definecolor{bgcolor}{rgb}{0.95,0.95,0.92}
\definecolor{spblue}{HTML}{00b5ea}

\lstdefinestyle{python}{
    language=Python,
    basicstyle=\fontsize{8}{10}\ttfamily,
    keywordstyle=\color{blue},
    commentstyle=\color{gray},
    stringstyle=\color{black},
    showstringspaces=false,
    breaklines=true,
    breakindent=0pt,
    breakatwhitespace=false,
    escapeinside={(*@}{@*)}
}

\lstdefinestyle{cpp}{
    language=C++,
    basicstyle=\fontsize{8}{10}\ttfamily,
    keywordstyle=\color{blue},
    commentstyle=\color{gray},
    stringstyle=\color{green},
    showstringspaces=false,
    breaklines=true,
    breakindent=0pt,
    breakatwhitespace=false,
    escapeinside={(*@}{@*)}
}

\lstdefinestyle{plain}{
    basicstyle=\fontsize{8}{10}\ttfamily,
    keywordstyle=\color{blue},
    commentstyle=\color{gray},
    stringstyle=\color{green},
    showstringspaces=false,
    breaklines=true,
    breakatwhitespace=false,
    breakindent=0pt,
    escapeinside={(*@}{@*)},
    literate={á}{{\'a}}1 {ã}{{\~a}}1 {é}{{\'e}}1,
}

\lstdefinestyle{demo}{
    basicstyle=\fontsize{8}{9}\ttfamily,
    keywordstyle=\color{blue},
    commentstyle=\color{gray},
    stringstyle=\color{green},
    showstringspaces=false,
    breaklines=true,
    breakatwhitespace=false,
    breakindent=0pt,
    escapeinside={(*@}{@*)},
    literate={á}{{\'a}}1 {ã}{{\~a}}1 {é}{{\'e}}1,
}

\lstdefinestyle{example}{
    basicstyle=\fontsize{8}{10}\ttfamily,
    keywordstyle=\color{spblue}\bfseries\underline,
    commentstyle=\color{gray},
    stringstyle=\color{green},
    showstringspaces=false,
    breaklines=true,
    breakatwhitespace=false,
    breakindent=0pt,
    escapeinside={(*@}{@*)},
    morekeywords={ Question, Answer, Prediction, Results, Explanation },
}

\lstdefinestyle{python2}{
    language=Python,
    basicstyle=\fontsize{8}{10}\ttfamily,
    keywordstyle=\color{blue},
    commentstyle=\color{gray},
    stringstyle=\color{green},
    showstringspaces=false,
    breakatwhitespace=false,
    breaklines=true,
    breakindent=0pt,
    escapeinside={(*@}{@*)}
}

\lstdefinestyle{cpp2}{
    language=C++,
    basicstyle=\fontsize{8}{10}\ttfamily,
    keywordstyle=\color{blue},
    commentstyle=\color{gray},
    stringstyle=\color{green},
    showstringspaces=false,
    breaklines=true,
    breakindent=0pt,
    breakatwhitespace=false,
    escapeinside={(*@}{@*)}
}

\lstdefinestyle{sql}{
    language=SQL,
    basicstyle=\fontsize{8}{10}\ttfamily,
    keywordstyle=\color{blue},
    commentstyle=\color{green},
    stringstyle=\color{black},
    showstringspaces=false,
    breakatwhitespace=false,
    breaklines=true,
    breakindent=0pt,
    escapeinside={(*@}{@*)}
}

\lstdefinestyle{prompt}{
    basicstyle=\fontsize{8}{10}\ttfamily,
    showstringspaces=false,
    breaklines=true,
    breakatwhitespace=false,
    breakindent=0pt,
    keepspaces=true,
    showspaces=false,  
    escapeinside={(*@}{@*)},
}

\lstdefinestyle{text}{
    basicstyle=\fontsize{8}{10}\ttfamily,
    showstringspaces=false,
    breaklines=true,
    breakatwhitespace=false,
    breakindent=0pt,
    keepspaces=true,
    showspaces=false,  
    escapeinside={(*@}{@*)},
}

\lstdefinestyle{defense-example}{
    basicstyle=\fontsize{8}{9}\,
    alsoletter={-},  %
    moredelim=[is][\color{cyan}\bfseries]{|}{|},  %
    keywordstyle=\color{red}\textsc\underline, %
    commentstyle=\color{gray},
    stringstyle=\color{green},
    showstringspaces=false,
    breaklines=true,
    breakatwhitespace=false,
    breakindent=0pt,
    escapeinside={(*@}{@*)},
    aboveskip=3pt, %
    belowskip=3pt,  %
    xleftmargin=1mm,
    xrightmargin=1mm,
}

\newcommand{\inserthl}[1]{\sethlcolor{promptinsert}\hl{#1}}
\newcommand{\comphl}[1]{\sethlcolor{compcolor}\hl{#1}}
\newcommand{\codehl}[1]{\sethlcolor{codehlcolor}\hl{#1}}
\newcommand{\codehlerr}[1]{\sethlcolor{codehlcolor2}\hl{#1}}

\newtcolorbox{mytitlebox}[2][]{ %
    enhanced,
    breakable,
    colbacktitle=gray!40,
    coltitle=black,
    title=#2,
    fonttitle=\small\scshape,
    after=\vspace{-3mm},  %
    boxed title style={size=small, colframe=black},
    #1 %
}

\lstdefinelanguage{mypython}{
  language=Python,
  basicstyle=\ttfamily\footnotesize,
  showstringspaces=false,
  breaklines=true,
  tabsize=2
}

\newtcolorbox[auto counter, list inside=mytextboxlist]{mytextbox}[2][]{%
  colback=gray!0,
  colframe=black,
  boxrule=0.5pt,
  left=3pt, right=3pt, top=3pt, bottom=3pt,
  arc=3pt,
  fontupper=\ttfamily\fontsize{8pt}{9pt}\selectfont,
  breakable,
  enhanced,
  listing only,
  listing options={basicstyle=\ttfamily\small},
  title={Listing~\thetcbcounter: #2}, %
  label type=mytextbox,               %
  list entry={#2},                    %
  #1
}

\newtcolorbox{questionbox}[1][]{
  enhanced,
  breakable,
  width=\linewidth,
  colback=green!12, colframe=green!50!black,
  colback=black!8, colframe=black!40,
  colbacktitle=black!40, coltitle=white, %
  title=Question,
  boxrule=0.6pt,
  arc=4pt, outer arc=4pt,
  left=8pt, right=8pt, top=6pt, bottom=6pt,
  fontupper=\ttfamily\small,
  #1
}

\newtcblisting{responsebox}[2][]{%
  enhanced,
  breakable,
  width=\linewidth,
  listing engine=listings,
  listing only,
  colback=black!4, colframe=black!40,
  colbacktitle=black!70, coltitle=white, %
  boxrule=0.5pt,
  arc=4pt, outer arc=4pt,
  left=8pt, right=8pt, top=6pt, bottom=6pt,
  title=#2~Response,
  listing options={
    language=mypython,
    basicstyle=\ttfamily\small,
    breaklines=true,
    breakatwhitespace=true,
    columns=fullflexible,
    keepspaces=true,
    showstringspaces=false
  },
  #1
}

\maketitle

\begin{abstract}

Large language models (LLMs) with integrated search tools show strong promise in open-domain question answering (QA), yet they often struggle to produce complete answer set to complex questions such as \textit{``Which actor from the film Heat won at least one Academy Award?''}, which requires \ding{182} distinguishing between multiple films sharing the same title and \ding{183} reasoning across a large set of actors to gather and integrate evidence.
Existing QA benchmarks rarely evaluate both challenges jointly. To address this, we introduce \datagen, an automatic data generation pipeline that constructs QA tasks grounded in text corpora and linked knowledge graph, generating natural and verifiable questions that systematically embed name ambiguity and multi-step reasoning. Based on this, we build \dataname, a dataset of 3,600 questions requiring multi-hop reasoning and half of them explicit name ambiguity resolving.
Experiments reveal that, even state-of-the-art GPT-5 show incomplete answers, achieving only 0.13 exact match on ambiguous questions and 0.21 on non-ambiguous questions. These findings highlight the need for more robust QA systems aimed at information gathering and answer completeness.
\end{abstract}
\section{Introduction}\label{sec:intro}

Large language models (LLMs) are increasingly used for complex question answering (QA), especially when paired with external search tools. Given an open-ended query, such systems decompose it into sub-steps, query the search tool, retrieve evidence, and aggregate information from multiple sources to form the final answer~\cite{openai2025deepresearch, jin2025searchr1, team2025kimik2}. Unlike simple factoid QA, which often requires a single lookup, complex queries are inherently multi-hop and demand multiple rounds of reasoning and evidence gathering~\cite{yang2018hotpotqa, trivedi2022musique, wolfson2025monaco}.

\begin{figure}[t]
    \centering
    \includegraphics[width=1.0\linewidth]{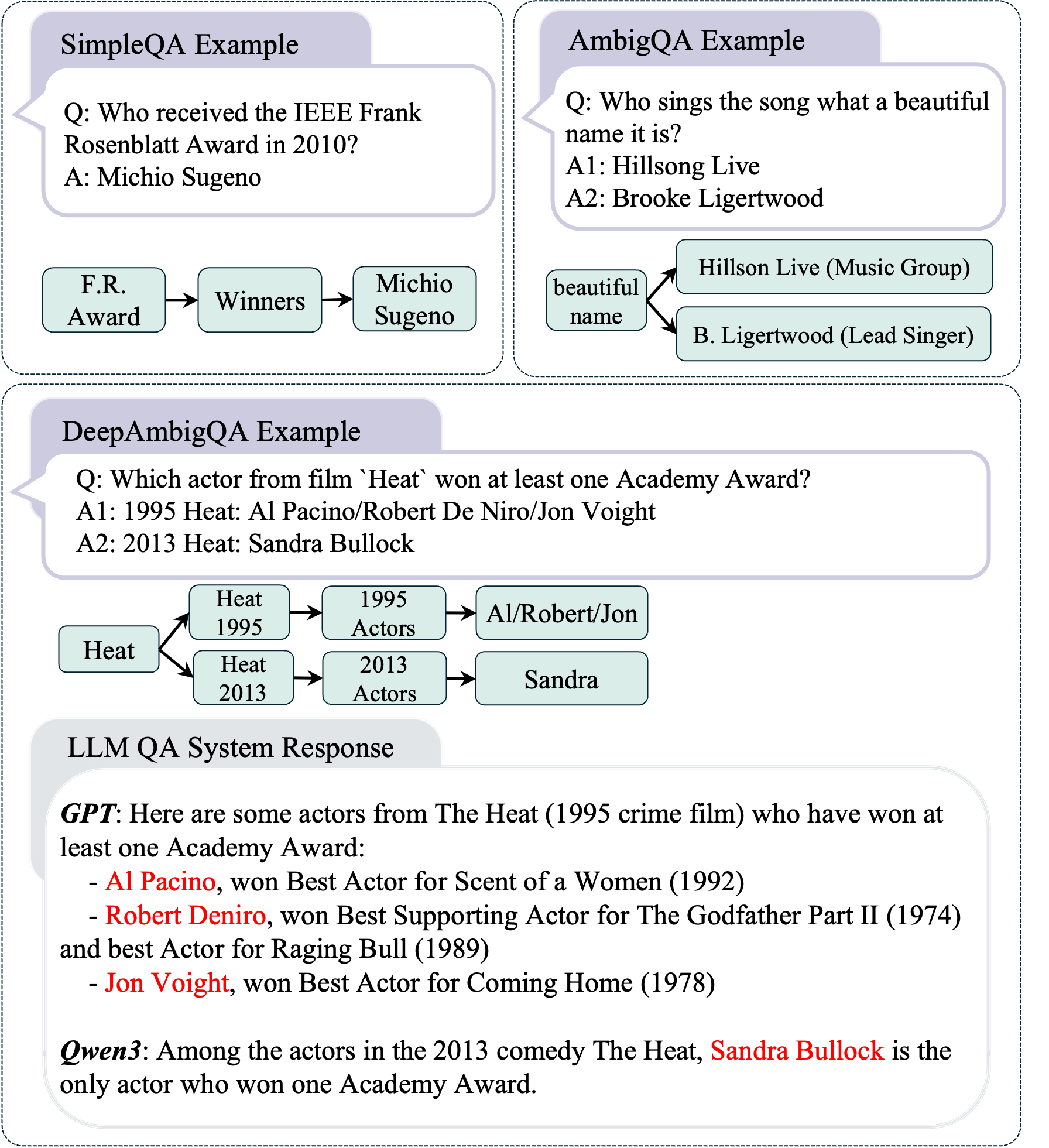}
    \vspace*{-5mm}
    \caption{Comparison between \dataname questions (bottom), SimpleQA (top left), and AmbigQA (top right). \textbf{\textit{Takeaway:}} Unlike SimpleQA and AmbigQA, \dataname questions require both name disambiguation and multi-hop reasoning over a large intermediate evidence set, presenting a significant challenge for current LLM-based QA systems.}
    \label{fig:intro-example}
    \vspace*{-0.3in}
\end{figure}

Two core capabilities are crucial for QA systems to solve such questions: 
\ding{182} \textit{Complete evidence coverage i.e.}, they must retrieve all relevant facts by issuing effective queries, disambiguating ambiguous inputs, and gathering the full set of related information rather than relying on a few salient snippets.
\ding{183} \textit{Correct reasoning and information aggregation i.e.}, they must then reason accurately over the collected evidence and perform reasoning operations such as filtering and grouping, while preserving contextual nuances.
For example, the query: \textit{``Which actors in the film titled The Heat have won at least one Academy Award?''} illustrates both challenges: the title \textit{The Heat} is ambiguous, as multiple films share this name, and solving the query requires enumerating all candidate films, retrieving their casts, and verifying each actor's award status. Thus, it demands both robust disambiguation and multi-hop reasoning over complete evidence. However, existing LLM-based QA systems often fall short, frequently overlooking valid interpretations or producing incomplete actor lists as shown in Figure~\ref{fig:intro-example}.

Although such queries are common, they remain underrepresented in existing QA benchmarks. Multi-hop QA datasets such as MuSiQue and SimpleQA~\cite{trivedi2022musique, wei2024simpleqa} generally follow a single reasoning path and produce single short answer. In contrast, ambiguity-oriented datasets such as AmbigQA and CondAmbigQA~\cite{min2020ambigqa, li2025condambigqa} capture multiple valid interpretations but involve limited multi-hop reasoning. Figure~\ref{fig:intro-example} illustrates how our target questions differ in both reasoning complexity and name ambiguity from those in prior benchmarks.
To address this gap, we introduce \datagen, an automatic pipeline for generating ambiguous multi-hop questions. Unlike prior QA dataset construction efforts that rely on manual annotation~\cite{trivedi2022musique, wolfson2025monaco},
\datagen\ leverages knowledge graphs aligned with a text corpora such as Wikidata and Wikipedia, which provide structured relations among entities that can be directly converted into executable reasoning steps and verified against factual connections. 
\datagen\ first identifies ambiguous names and their corresponding distinct entities, then builds executable reasoning plans with structured sequences of operations that model user information-seeking behaviors. These plans simulate multi-step reasoning by chaining operations such as \textit{entity} retrieval (\textit{e.g.}, obtaining actors of a film) and \textit{filter} application (\textit{e.g.}, selecting actors who have won an award). 
For ambiguous names, the same plan can be re-executed with each possible entity to obtain corresponding answers, allowing the dataset to capture multiple valid interpretations in a verifiable way. Finally, an LLM converts these plans into natural-language questions. We also integrates verification steps to ensure naturalness and question-answer correspondence.

Building on \datagen, we introduce \dataname, a benchmark of 3,600 QA pairs. Among them, 1,800 questions contain ambiguous names leading to multiple valid reasoning paths, while the rest are non-ambiguous but still yield large answer sets (up to 43 entities). Each question requires at least two reasoning steps, with some involving up to eight, posing substantial challenges for LLM QA systems on complete answer recall and intermediate reasoning. 
We evaluate state-of-the-art LLMs equipped dense retrieval tool on Wikipedia corpus on \dataname. The results show that:
\ding{182} LLMs struggle with answer completeness, achieving low exact-match accuracy, especially for ambiguous queries, even the SOTA GPT-5 attains only 0.13 exact match on the ambiguous subset and 0.21 exact match on the non-ambiguous subset.
\ding{183} While precision remains high (typically above 70\%), recall and exact match are much lower, indicating their inability to find complete answers. Adding modules such as query expansion for disambiguation and evidence extraction for retrieved passage brings only modest improvements and cannot fully address these challenging questions.

In summary, our contribution are as follows:
\begin{itemize}\setlength{\itemsep}{0pt}
\item We propose \datagen, an automatic QA synthesis pipeline grounded in knowledge graphs that generates natural questions and verifiable gold answers.
\item We introduce \dataname, a benchmark of multi-hop, ambiguity-rich questions requiring reasoning and disambiguation to ensure answer completeness for QA systems.
\item We evaluate SOTA LLMs and demonstrate that they struggle with answer recall and completeness, highlighting the need for more robust QA systems.
\end{itemize}

\section{Related Work}\label{sec:related-work}

\begin{figure*}[t]
    \centering
    \includegraphics[width=0.95\linewidth]{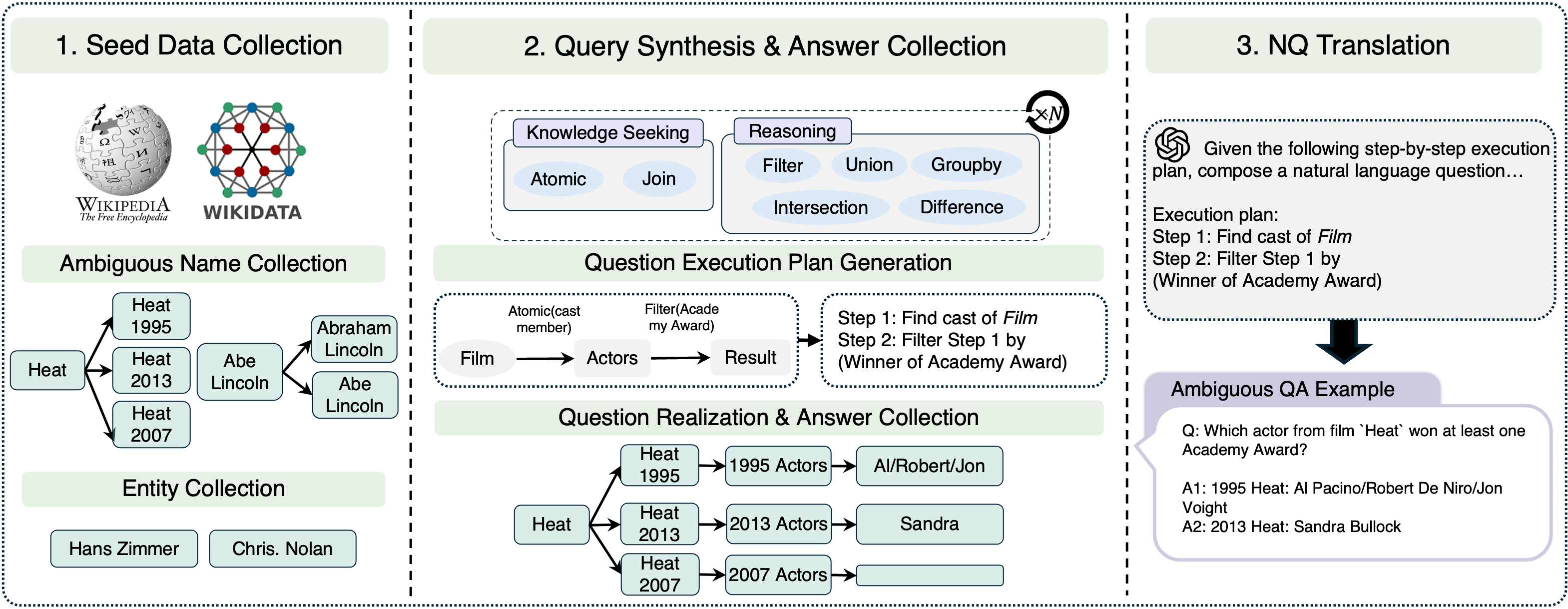}
    \vspace*{-2mm}
    \caption{
    Illustration of the \datagen algorithm. Beginning with a large collection of entities and relations, a complex query is constructed by iteratively sampling atomic entities and new operations. The resulting query is then converted into natural language by prompting an LLM.}
    \label{fig:datagen-algo}
\end{figure*}
\paragraph{Ambiguous QA}
Ambiguous QA datasets focus on questions that can be interpreted in multiple valid ways. Existing benchmarks capture different types of ambiguity: AmbigQA, ASQA, and CondAmbigQA mainly address name entity and intent ambiguities~\cite{min2020ambigqa, stelmakh2022asqa, li2025condambigqa}, while TempAmbigQA targets temporal ambiguity~\cite{piryani2024tempambiqa}. 
Although these datasets cover diverse ambiguity types, most of their questions require only shallow evidence collection without deep, multi-step reasoning.
In contrast, our QA tasks involve complex, multi-step reasoning and information aggregation for various name entity interpretations in the question, which substantially increases the difficulty and richness of the questions.

\paragraph{Multi-hop QA}
Multi-hop QA datasets typically contains questions requiring strong reasoning and evidence aggregation across documents.
Datasets such as HotpotQA, 2WikiMultihopQA, and MuSiQue study reasoning over a limited number of documents and a small set of structured steps~\cite{yang2018hotpotqa, ho2020twikimultihopqa, trivedi2022musique}. More recent datasets, like FanOutQA, MEQA, WebQuest, and MoNaCo, further raise the difficulty by expanding to larger document pools, event-centric compositions, and long-horizon reasoning sequences~\cite{zhu2024fanoutqa,li2024meqa,wang2024webquest,wolfson2025monaco}.
However, most of these assume a \emph{single reasoning path and a single final answer} for the question, leaving ambiguity underexplored. 
\dataname\ differs by introducing entity disambiguation that creates multiple valid branches and focusing more on the answer completeness across all branches. 
Unlike prior datasets that rely on human annotation, which is very difficult for these multi-hop reasoning questions, our QA construction can be operated automatically, largely reducing human-annotation cost.

\paragraph{Knowledge-graph QA}
KG QA datasets use knowledge graphs as the information source for QA synthesis. For example, \text{WebQuestionsSP} pairs natural questions with executable semantic parses over Freebase~\cite{yih2016webqsp}, while \text{ComplexWebQuestions} composes multi-hop queries grounded in web evidence~\cite{talmor2018cwq}. \text{GrailQA} targets compositional and schema-level generalization~\cite{gu2021grailqa}, and program-supervised datasets such as \text{MetaQA}, \text{CFQ}, and \text{KQA~Pro} provide fine-grained logical forms for controlled reasoning~\cite{zhang2018metaqa,keysers2020cfq,cao2022kqapro}. Recent works like \text{MusTQ} and \text{Dynamic-KGQA} also introduce temporal and adaptive reasoning~\cite{zhang2024mustq,dammu2025dynamickgqa}. Yet most KGQA tasks still assume a single executable program, limiting evaluation to single reasoning path. Our approach instead generates KG-grounded questions aligned with Wikipedia, enabling complex questions that require multi-hop reasoning across ambiguous name entities.

\begin{table*}[t]
\centering
\vspace{3mm}
\small
\resizebox{\textwidth}{!}{
\begin{tabular}{p{5.3cm} p{2.5cm} p{5.0cm} p{4.5cm}}
\toprule \midrule
\textbf{Synthesized Query} & \textbf{Atomic Entities} & \textbf{Question Execution Plans} & \textbf{Final Answers} \\
\midrule
Which directors have worked with Saoirse Ronan on two or more films?
& Saoirse Ronan 
& (1) Find film by Saoirse Ronan. (2) Find director of step 1 result. (3) Group by director. (4) Filter by count > 2.
& \{Joe Wright, Greta Gerwig, Wes Anderson\}
\\
List all actors who have starred in both a Martin Scorsese film and a Quentin Tarantino film.
& Martin Scorsese, Quentin Tarantino
& (1) Find films by Scorsese. (2) Join step 1 result with cast member. (3) Find films by Tarantino. (4) Join step 3 result with cast member. (5) Intersection on step 2 and step 4.
& \{Robert De Niro, Leonardo DiCaprio, Harvey Keitel, Jonah Hill, Brad Pitt, Margot Robbie, Ray Liotta\}
\\

Which films were directed by one of the cast members of City of God? & City of God (2002 film), City of God (2011 film) & (1) Find cast of City of God (2) Join step (1) result with directed film & 2002 film: \{City of God 2013 10 Years Later, The Dead Girl's Feast\}, 2011 film: \{Appavin Meesai, Lucifer, Bro Daddy, L2: Empuraan\}
\\
Which genres are shared by Americano and Sleepless in Seattle? & Americano (2011 film), The Americano (1916 film) & (1) Find genres of Americano (2) Find genres of Sleepless in Seattle (3) Intersection of Step 1 and 2 & 1916 film: \{drama film\}, 2011 film: \{romantic comedy\} 
\\

\midrule
\bottomrule
\end{tabular}
}
\caption{Example questions generated by \datagen, along with the involved entities in question, execution plans, and final answers. Row 3 and 4 show examples of ambiguous question, where the ambiguity comes from the shared film name.}
\label{tab:example-query}
\end{table*}

\section{\datagen: An Automatic QA Synthesis Pipeline}\label{sec:datagen}
\subsection{Pipeline Overview}\label{subsec:datagen-overview}
In this section, we present \datagen, an automatic pipeline for synthesizing ambiguous multi-hop knowledge-seeking QAs. The pipeline is designed to capture diverse reasoning patterns, and explicitly model name entity ambiguity that induce multiple reasoning paths. \datagen involves three stages:
\ding{182} \textit{Seed Data Collection.} The pipeline begins with the collection of seed entities and relations. These constitute the atomic building blocks for multi-hop reasoning and serve as the foundation for question synthesis.  
\ding{183} \textit{Execution Plan Synthesis and Answer Collection.} \datagen utilizes the seed data and composes structured execution plans that specify multi-step reasoning operations. These plans can be executed on a knowledge graph to yield ground-truth answers. KG ensures the correctness and completeness of the answers.
\ding{184} \textit{Natural Language Question Translation and Filtering.}  Execution plans are translated into natural language questions with an LLM, \textit{i.e.}, GPT-5-mini. The candidate QAs are further filtered to ensure semantic naturalness and question-answer correspondence. Figure~\ref{fig:datagen-algo} presents a pipeline overview.

\subsection{\datagen Algorithm}
\label{subsec:datagen-algo}

\paragraph{Step 1: Seed Data Collection}
The pipeline begins by defining a domain schema consisting of entity types and relations that later guide information-seeking and reasoning. For example, in the film domain, entity types can include \textit{actors}, \textit{directors}, and \textit{films}, while relations capture the semantic transition between them such as \textit{directed by} (director-film) or \textit{cast member} (actor-film). This schema specifies the space of valid reasoning paths and constrains how knowledge can be retrieved or composed. We derive the schema from Wikidata KG definitions to ensure interpretability and consistency with the Wikipedia text corpus, and obtain concrete entity instances for each type via WikiSPARQL queries.

\paragraph{Step 2: Query Synthesis and Answer Collection}
Using the seed data, the pipeline incrementally builds executable plans that represent typical multi-step reasoning and information-seeking behaviors of complex user questions. Each plan consists of a sequence of operations sampled from seven atomic types: \texttt{Atomic}, \texttt{Join}, \texttt{Filter}, \texttt{Union}, \texttt{Difference}, \texttt{Intersection}, and \texttt{GroupBy}.

The first two operations, \texttt{Atomic} and \texttt{Join}, capture knowledge-seeking behaviors. Specifically, \texttt{Atomic} starts a traversal from a single entity through a specified relation, while \texttt{Join} extends the traversal from an intermediate set of entities to gather related entities. For example, \texttt{Atomic(Tom Hanks, cast member)} retrieves the set of films featuring Tom Hanks, and \texttt{Join(Hanks’ films, directed by)} returns the directors of those films.

The remaining operations model reasoning behaviors including applying logical constraints, performing set-based computations, or aggregating results. These include filtering answer set based on certain predicates such as award-winning status, computing unions, intersections, or differences between intermediate answer sets, and grouping current answer entities for aggregation. 
We highlight that all operations are fully executable in code, allowing consistent re-instantiation and execution across different entities, which guarantees the answer correctness and completeness.
It further helps handle ambiguous questions, where we can re-run the same question execution plan on each unique entities sharing same ambiguous name to derive the unique answer sets for each reasoning branch. The mid panel of Figure~\ref{fig:datagen-algo} shows an example that executing the plan yield three different answer sets for three different Heat film entity.

\paragraph{Step 3: Natural Language Question Translation}
After generating an execution plan and collecting its answers, the pipeline uses a large language model (LLM) to translate the formal representation into a natural language question. The translation process is guided by both the execution plan and its final answer set, ensuring that the resulting question reflects the underlying reasoning steps. Candidate questions are further filtered to remove those whose semantics diverge from the plan or whose answers do not align. 
As shown in Table~\ref{tab:example-query}, this process produces natural language questions that capture complex, multi-hop reasoning patterns while remaining answerable.

\subsection{Additional Details}\label{subsec:datagen-detail}

\paragraph{QA Filtering}
To ensure that the generated questions and answers are coherent and answerable, the pipeline applies two complementary filtering strategies.  
\ding{182} \textit{Heuristic rules.} These rules enforce syntactic and semantic validity. For example, we introduce rules about preventing repetitive reasoning operations (\textit{e.g.}, consecutive set unions), disallowing incompatible combinations of entity types (\textit{e.g.} union between actor and film), and remove cases where the final answer set is empty.  
\ding{183} \textit{LLM-based evaluation.} Since heuristics alone cannot fully cover unreasonable question semantics, we prompt GPT-5-mini to further assess both the logical validity of execution plans and evaluate the alignment between generated questions and resulting answer sets. This helps us remove meaningless and unnatural questions.

\paragraph{LLM-based Entity Selection}
Although many entity candidates can be used in an execution plan, most do not yield valid answer sets due to filtering or set intersection. To improve the synthesis success rate, we employ an LLM-based entity realization strategy. Specifically, we randomly sample a subset of candidate entities corresponding to the entity types in an execution plan from the seed entity sets, \textit{e.g.} 50 different films for an execution plan requiring a film, and prompt GPT-5-nano to select suitable entities for realization that are likely to produce non-empty answer sets. This approach substantially improves the success rate of generating valid QA pairs with non-empty answers.

\section{\dataname Dataset}\label{sec:dataset}

In this section, we present more details of the \dataname dataset, generated by the proposed \datagen on a Wikipedia snapshot\footnote{\url{https://en.wikipedia.org/wiki/Wikipedia:Database_download}}. 
Specifically, Section~\ref{subsec:data-collection} details the data collection procedure, Section~\ref{subsec:data-stats} lists the key statistics of \dataname dataset, followed by Section~\ref{subsec:data-analysis} that presents the complexity and naturalness analysis of the synthesized questions.

\subsection{\dataname Collection}\label{subsec:data-collection}
In this paper, we employ the Wikipedia snapshot dated \textit{2025-0820}\footnote{\url{https://dumps.wikimedia.org/enwiki/20250820/}} as the source data to build \dataname. The wiki pages are aligned to corresponding wikidata annotations to obtain the knowledge graph with entities and relation annotations for query synthesis.

The final dataset contains both ambiguous and non-ambiguous questions spanning a diverse set of domains derived from Wikipedia, including \textit{movies, music, sports, books, business, and science}. 
To better capture real-world query ambiguity, we collect the ambiguous name as well as the unique entities corresponding to the same name from Wikipedia disambiguation page, \textit{e.g.} the page \textit{Heat (disambiguation)}\footnote{\url{https://en.wikipedia.org/wiki/Heat_(disambiguation)}} collects unique entities from multiple domains sharing the same ambiguous name \textit{Heat}.

\begin{figure}[t]
  \resizebox{0.95\linewidth}{!}{
    \begin{tabular}{p{0.9\linewidth}r}
      \toprule
      \midrule
      \bf \textbf{Non-ambiguous questions}: & 1800 \\
      \quad - Avg. \# question words &  13.9 \\
      \quad - Avg. \# answer set size & 8.8 \\
      \quad - Avg. \# execution steps & 4.3 \\
      \quad - Avg. \# entities involved & 463.1 \\
      \midrule 
      \bf \textbf{Ambiguous questions}: & 1800 \\ 
      \quad - Avg. \# question words & 12.4 \\
      \quad - Avg. \# answer set size & 10.3 \\
      \quad - Avg. \# execution steps & 3.7 \\
      \quad - Avg. \# answer set per branch & 2.7 \\
      \quad - Avg. \# entities involved & 247.4 \\
      \quad - Avg. \# reasoning branches & 3.3 \\
      \midrule
      \bottomrule
    \end{tabular}
  }
  \vspace*{-2mm}
  \captionof{table}{Key data statistics of \dataname.}
  \label{tab:data-stats}
  \vspace*{-0.1in}
\end{figure}
\begin{figure}[t]
  \centering
  \resizebox{0.95\linewidth}{!}{
    \begin{tabular}{lccc}
      \toprule \midrule
      \textbf{Metric} & \textbf{Non-Ambig. Question} & \textbf{Ambig. Question} \\
      \midrule
      Naturalness(1-3) & 2.86 & 2.91 \\
      Correctness(\%) & 91.6 & 91.2 \\
      \midrule
      \bottomrule
    \end{tabular}
  }
  \captionof{table}{ Human evaluation results of \dataname on question naturalness and answer correctness. Scores are averaged over three annotators. }
  \label{tab:human-eval}
\end{figure}
\subsection{Dataset Statistics}\label{subsec:data-stats}
Table~\ref{tab:data-stats} summarizes the key characteristics of \dataname, divided into ambiguous question subset and non-ambiguous question subset. For each category, we report the average question length, total answer set size, number of execution steps, and number of unique intermediate entities involved in the reasoning, which together reflect the overall complexity of the questions. Detailed breakdown such as histogram of the execution step and answer set size are in Appendix~\ref{subsec:app-data-detail}.

The ambiguous subset contains 1,800 questions, each involving one ambiguous name that yield multiple interpretations.
Such questions introduce substantial challenges for LLMs to perform accurate disambiguation and retrieval across multiple reasoning paths starting from different entities.
On average, an ambiguous question has 3.3 reasoning branches and involves 247.4 entities in reasoning process, underscoring the difficulty of resolving entity ambiguities within the reasoning process.
The non-ambiguous subset also includes 1,800 questions but emphasizes more on reasoning complexity.
These questions typically involve deeper reasoning chains, averaging 4.3 execution steps and covering 463.1 unique entities.
They demand compositional reasoning operations such as filtering, set intersection, and aggregation, which test an LLM’s ability to maintain logical consistency over extended reasoning paths.

Overall, \dataname captures two complementary sources of reasoning difficulty: ambiguity resolution and compositional reasoning.
Ambiguous questions pose more challenges on QA system's capacity to differentiate among entities with identical names, while non-ambiguous questions probe its ability to perform multi-step reasoning over complex relational structures.
Together, these characteristics make \dataname\ a challenging benchmark for evaluating the robustness and reasoning depth of QA systems.

\subsection{Dataset Quality Analysis}\label{subsec:data-analysis}
We further conduct a human study to better assess the quality of \dataname. 
The primary objective is to evaluate whether the synthesized questions are natural in phrasing and whether their corresponding answers are correct and faithful, ensuring that the dataset reflects realistic user queries while maintaining logical soundness. 
For each annotation task, we recruit 3 human annotators and report the average. Due to budget constraints, we randomly select 5\% of the non-ambiguous and ambiguous QAs, with 90 unique questions for each subset. 
Detailed annotation instructions are provided in Appendix~\ref{subsec:app-humaneval}.

\paragraph{Human Study on Naturalness.}  
We evaluate the naturalness of synthesized questions to determine whether they are grammatically correct, clearly capture the intended knowledge search task, and are phrased in a way that is natural and easily understandable to humans. Human annotators rated each question on a 3-point scale, yielding a pairwise Cohen's Kappa of 0.669 and 0.632 on the non-ambiguous and ambiguous subsets, implying a substantial agreement. As shown in Table~\ref{tab:human-eval}, the synthesized questions received high naturalness scores, averaging 2.86 for non-ambiguous questions and 2.91 for ambiguous ones.

\paragraph{Human Study on Correctness.}
We additionally asked annotators to assess the correctness of answers by verifying them against the corresponding Wikipedia pages and Wikidata entries. The goal was to ensure that answers are both accurate and complete with respect to the original questions. 
For ambiguous questions, we ask annotators to evaluate the correctness of unique entities for the ambiguous name in original question to avoid their burden.
As shown in Table~\ref{tab:human-eval}, we achieve a high entity-level correctness score, with 91.6\% for the non-ambiguous subset and 91.2\% for the ambiguous subset, with pair-wise Cohen's Kappa of 0.774 and 0.534 respectively. The lower coefficient for the ambiguous subset (moderate agreement) also highlights the difficulty in verifying the correctness of the answer for ambiguous questions.

\section{Experiment}\label{sec:experiment}

In this section, we evaluate a diverse set of SOTA LLMs on \dataname\ dataset and further investigate the effect of additional modules in the system on their overall performance.
Section~\ref{subsec:exp-setting} describes the experimental setup and presents quantitative results across different model families, accompanied by a detailed analysis of failure cases in Section~\ref{subsec:exp-result}.
Finally, Section~\ref{subsec:exp-analysis} examines the contribution of each add-on module and its influence on the final performance.

\subsection{Experiment Setting}\label{subsec:exp-setting}
\paragraph{Evaluation Model}
We evaluate 10 large language models (LLMs) spanning both closed-source and open-source families, including the GPT series~\cite{openai2023gpt4}, Qwen2.5-Instruct series~\cite{yang2025qwen2_5}, Qwen3 series~\cite{yang2025qwen3}, and the DeepSeek-R1-distilled series~\cite{deepseekai2025deepseekr1incentivizingreasoningcapability}. All models are prompted with the same template to summarize their answers in a JSON dictionary format to simplify answer extraction. 
For ambiguous questions that may involve multiple branches and different answer sets, we incorpoarate an LLM-based matching step to align the LLM generated answers with the ground-truth branch names. 
Implementation details including the detailed prompts are in Appendix~\ref{sec:app-implementation} and ~\ref{subsec:app-prompt}.

Following previous work~\cite{yifei2025researchqa, jin2025searchr1}, we implement LLM-based QA system by integrating a dense-retrieval search tool, and prompt LLMs to query the local Wikipedia corpus via the provided search tool for information seeking. 
In this process, the LLM may trigger multiple tool calls to retrieve new information that is later appended to their context.
The Wikipedia corpus is preprocessed and served following the FlashRAG~\cite{FlashRAG} implementation. Unless otherwise specified, we employ e5-base\footnote{\url{https://huggingface.co/intfloat/e5-base-v2}} as the embedding model for corpus indexing and retrieval following previous work~\cite{jin2025searchr1, mihaylov2021conditional}. 
More implementation details are in Appendix~\ref{sec:app-implementation}.

\paragraph{Evaluation Metric}
Since all answers in \dataname\ are represented as sets of entities, we evaluate model performance using average precision (P), recall (R), and exact match (EM).
For ambiguous questions, results are aggregated over all valid answers spanning different reasoning branches to account for multiple interpretations.

\subsection{Experiment Results}\label{subsec:exp-result}

\begin{table}[t]
\centering
\renewcommand{\arraystretch}{1.2}
\resizebox{1.0\linewidth}{!}{%
\begin{tabular}{l|cccccc}
\toprule \midrule
\multirow{2}{*}{Model} & 
\multicolumn{3}{c}{\textbf{Non-ambig. Question}} & 
\multicolumn{3}{c}{\textbf{Ambig. Question}} \\
\cmidrule(lr){2-4} \cmidrule(lr){5-7}
 & P & R & EM & P & R & EM \\
\midrule
Qwen2.5-3B-Inst       & 0.82 & 0.34 & 0.03 &  0.78 & 0.17 & 0.01 \\
Qwen2.5-7B-Inst         & 0.83 & 0.40 & 0.05 &  0.80 & 0.21 & 0.01 \\
Qwen3-4B                & 0.84 & 0.45 & 0.07 &  0.84 & 0.25 & 0.03 \\
Qwen3-8B                & 0.84 & 0.55 & 0.12 &  0.85 & 0.39 & 0.05 \\
DS-R1-distill-1.5B      & 0.73 & 0.33 & 0.02 &  0.74 & 0.08 & 0.01 \\
DS-R1-distill-7B        & 0.82 & 0.51 & 0.06 &  0.79 & 0.14 & 0.02 \\
GPT-4o                  & 0.84 & 0.52 & 0.11 &  0.83 & 0.33 & 0.03 \\
GPT-4o-mini             & 0.91 & 0.57 & 0.14 &  0.84 & 0.41 & 0.05 \\
GPT-5-mini              & 0.90 & 0.64 & 0.16 &  0.87 & 0.44 & 0.09 \\
GPT-5                   & \textbf{0.93} & \textbf{0.71} & \textbf{0.21} &  \textbf{0.91} & \textbf{0.48} & \textbf{0.13} \\
\midrule
\bottomrule
\end{tabular}
}
\vspace*{-3mm}
\caption{Evaluation results of various LLMs. }
\label{tab:main-result}
\vspace*{-0.2in}
\end{table}

\paragraph{LLM Performance on \dataname}
Table~\ref{tab:main-result} reports the overall performance of various LLMs on \dataname.
We highlight the following observations:
\ding{182} All LLMs exhibit low exact match scores, with the strongest model, GPT-5, achieving only 0.21 and 0.13 EM on the non-ambiguous and ambiguous subsets, respectively. This gap underscores that current LLMs struggle to maintain answer completeness when reasoning over complex questions that involve ambiguity resolution and intermediate information aggregation.
\ding{183} Although most models attain relatively high precision, their recall remains consistently lower.
This discrepancy suggests that LLMs often retrieve partial but not exhaustive sets of correct answer entities, revealing limitations in comprehensive information retrieval and reasoning coverage.
\ding{184} Performance degrades further on ambiguous questions. For instance, GPT-5’s recall drops from 0.71 to 0.48 when ambiguity is introduced, indicating that existing LLMs remain ineffective at disambiguating ambiguous names in question and exploring multiple valid reasoning branches.
Overall, these results demonstrate that \dataname\ presents substantial reasoning and disambiguation challenges to LLM QA systems.

\begin{figure}[t]
    \centering
    \includegraphics[width=\linewidth]{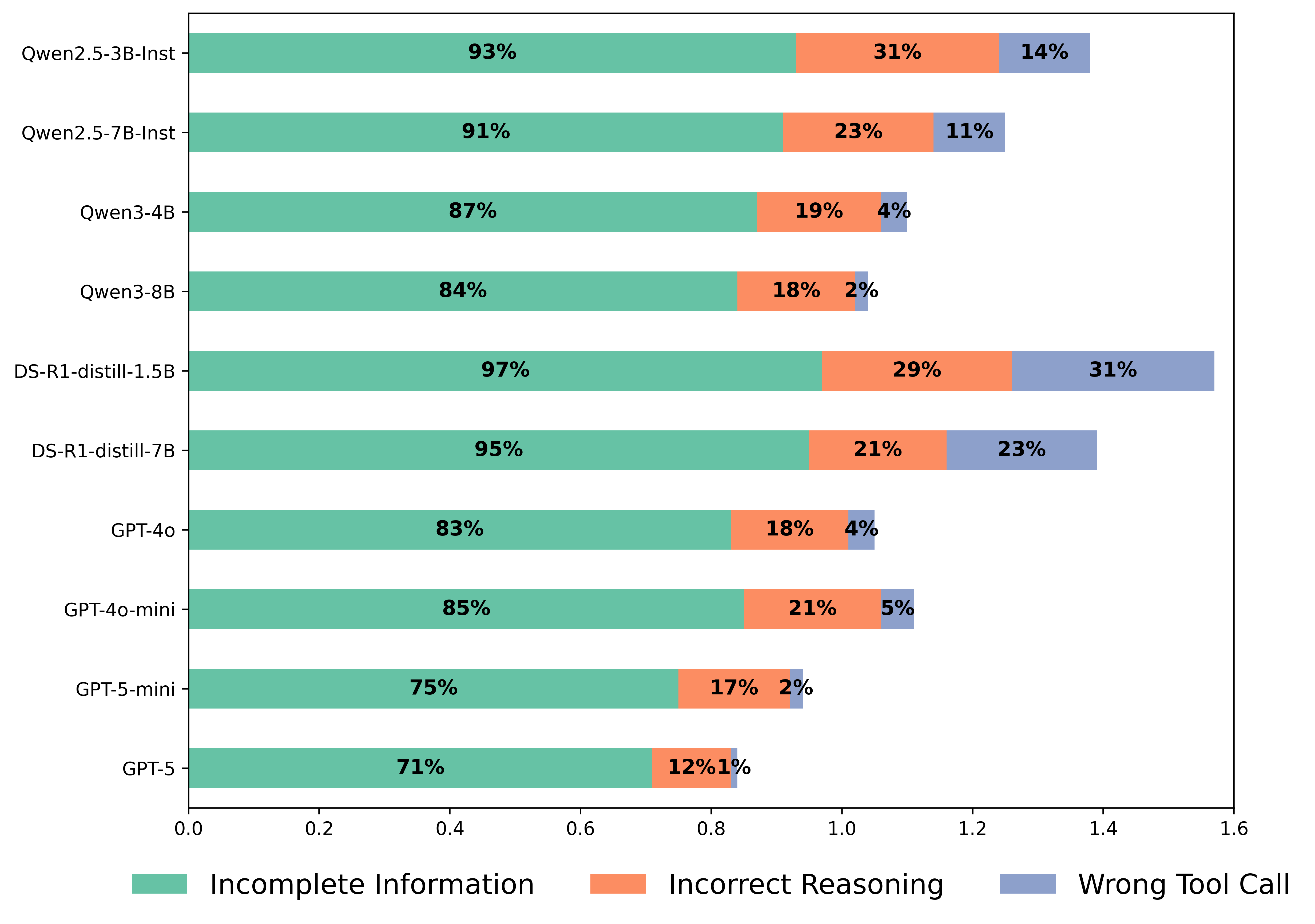}
    \vspace*{-7mm}
    \caption{Error-type breakdown for various LLMs.}
    \label{fig:failure-breakdown}
    \vspace*{-0.23in}
\end{figure}

\paragraph{Failure Breakdown and Analysis}
To better understand error cases in these LLM QA systems, we categorize errors into following three types: 
\ding{172} \textit{incomplete information extraction}, where answer entities are not involved in the retrieved documents. For ambiguous questions, this also includes the scenerio where certain reasoning branches is not covered.
\ding{173} \textit{incorrect intermediate reasoning}, where answer entities appeared in the retrieved documents but were not presented in LLM's final answer.
\ding{174} \textit{wrong tool call}, including cases of misused or failed tool execution.
Figure~\ref{fig:failure-breakdown} presents the distribution of these failure types. Note that the proportions in each bar do not sum to one because multiple error types can occur in a single QA instance. We highlight two main observations.
\ding{182} Stronger LLMs make fewer reasoning-related mistakes. For example, weaker LLMs such as Qwen2.5-3B exhibit over 30\% reasoning errors, whereas GPT-5 shows only 12\%.
\ding{183} All LLMs suffer from incomplete information extraction; even the strongest model, GPT-5, exhibits 71\% of failures in this category.
These findings suggest that enhancing retrieval information coverage remains the dominant challenge for LLM-based QA.

\subsection{Additional Analysis}\label{subsec:exp-analysis}
The previous results indicate that the base system still suffers from substantial retrieval and reasoning errors. To inform future improvements for QA systems, we explore three additional modules:
\ding{182} \textit{Query expansion}, which prompts the LLM to enrich original query and disambiguate before question answering to improve information completeness in reasoning process.
\ding{183} \textit{Evidence extraction}, which prompts the LLM to filter retrieved texts and retain only relevant spans to reduce information aggregation errors.
\ding{184} \textit{KG query}, a new tool that enables the LLM to execute WikiSPARQL queries on the KG, potentially providing precise entity-level information beyond textual evidence.
Implementation and full results are in Appendix~\ref{sec:app-more-exp}.

\begin{figure}[t]
    \centering
    \includegraphics[width=\linewidth]{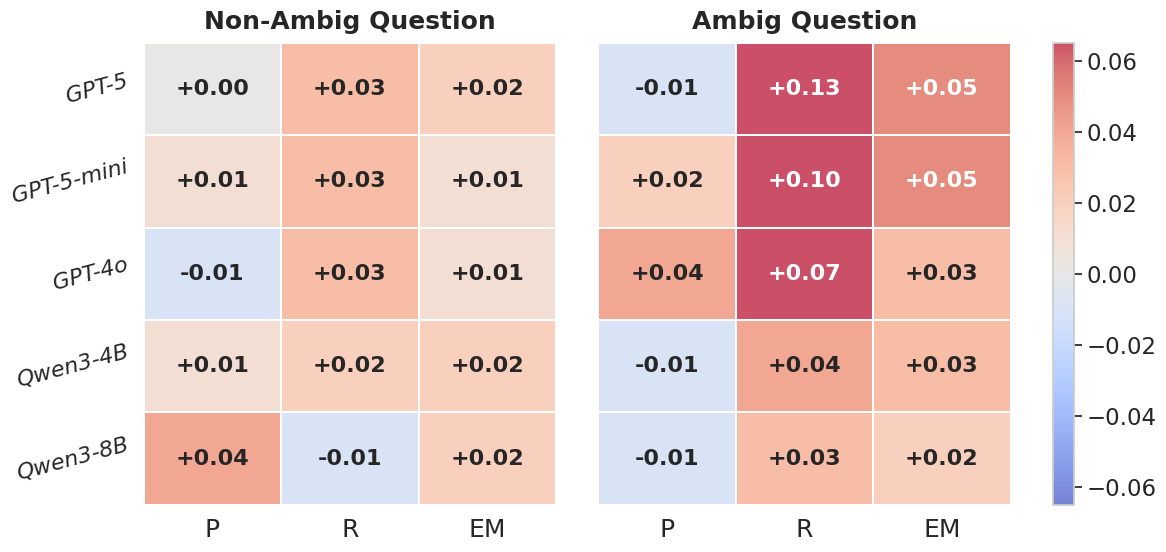}
    \includegraphics[width=\linewidth]{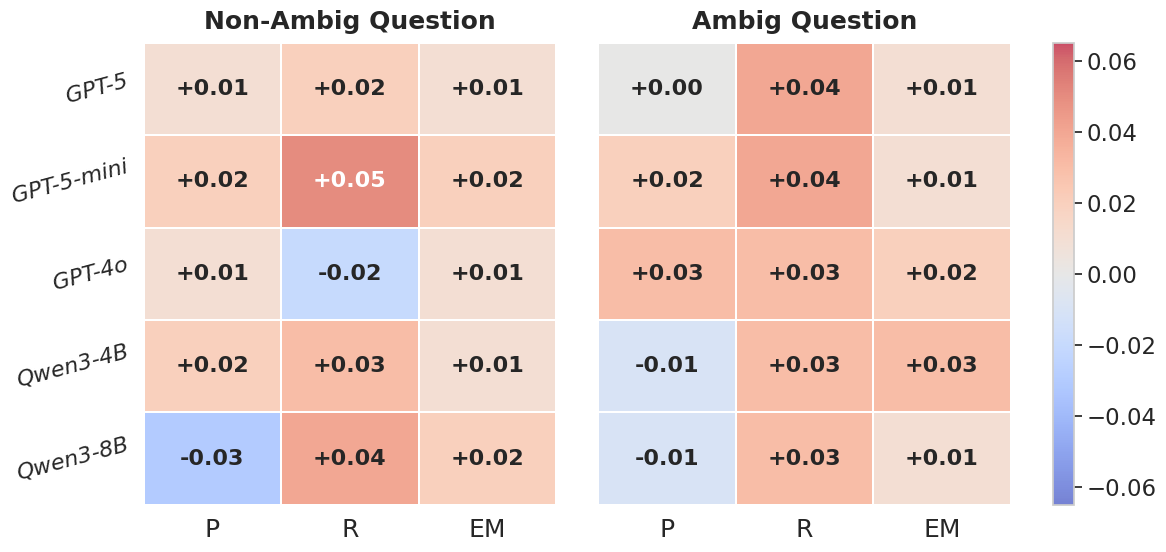}
    \caption{LLM performance change after applying query expansion \textit{(upper)} and evidence extraction \textit{(lower)}. Full results in Appendix Table~\ref{tab:module-ablation}.
    }
    \label{fig:ablation-extraction-expansion}
    \vspace*{-0.25in}
\end{figure}

\paragraph{Effect of Query Expansion}
Query expansion aims to resolve ambiguity in entity names and improve retrieval coverage by performing explict disambiguation step. As shown in Figure~\ref{fig:ablation-extraction-expansion}, adding expansion generally improves recall across all models, particularly for ambiguous questions. For instance, GPT-5’s recall shows a 0.13 increase on ambiguous queries, reflecting that expanded queries effectively helps LLM identify additional evidence that would otherwise be missed. Meanwhile, improvement is smaller for non-ambiguous questions, suggesting that it mainly benefits cases where entity names are unclear. 
However, precision and exact match gains are marginal, indicating that while the expanded queries help retrieve more relevant evidence, they still fail to fully address the answer completeness issue.

\paragraph{Effect of Evidence Extraction}
Evidence extraction module encourages the LLM to filter retrieved information and retain only those relevant to the question to reduce context noise overhead. As shown in Figure~\ref{fig:ablation-extraction-expansion}, this step yields modest gains in both precision and recall. For example, GPT-5 exhibits a 0.01 and 0.02 improvement for precision and recall on non-ambiguous subset. 
These small improvements indicate that extraction helps reduce irrelevant information but does not fully address cases where evidence is missing. 
This also aligns with our error analysis in Section~\ref{subsec:exp-result}, which shows that QA system suffer more from incomplete information rather than incorrect reasoning.

\paragraph{Effect of Knowledge-graph Information}
We also test whether using a Wiki-SPARQL tool to access KG can enhance model performance. This allows LLM to retrieve precise entity information rather than relying on text retrieval alone. However, Table~\ref{tab:module-ablation} shows a clear performance drop across all models. For instance, GPT-5 sees a notable 0.77 precision and 0.36 recall decline on ambiguous subset.
This drop mainly comes from the failure in generating correct and executable SPARQL queries such as incorrect grammar or wrong entity/relation id usage, limiting their improvement.  
The findings suggest that while KGs can offer structured access to entity relations, simply integrating them remains unreliable and require more advanced special handling.

\begin{table}[t]
\centering
\resizebox{1.0\linewidth}{!}{
\begin{tabular}{l|ccc|ccc}
\toprule \midrule
\multirow{2}{*}{Model} & 
\multicolumn{3}{c}{\textbf{Non-ambig. Question}} & 
\multicolumn{3}{c}{\textbf{Ambig. Question}} \\
\cmidrule(lr){2-4} \cmidrule(lr){5-7}
 & P & R & EM & P & R & EM \\
\midrule
GPT-5      & 0.30$_{\textcolor{red}{-0.63}}$ & 0.28$_{\textcolor{red}{-0.43}}$ & 0.03$_{\textcolor{red}{-0.18}}$ &
              0.14$_{\textcolor{red}{-0.77}}$ & 0.12$_{\textcolor{red}{-0.36}}$ & 0.02$_{\textcolor{red}{-0.11}}$ \\
GPT-5-mini & 0.25$_{\textcolor{red}{-0.65}}$ & 0.22$_{\textcolor{red}{-0.42}}$ & 0.02$_{\textcolor{red}{-0.14}}$ &
              0.11$_{\textcolor{red}{-0.76}}$ & 0.10$_{\textcolor{red}{-0.34}}$ & 0.01$_{\textcolor{red}{-0.08}}$ \\
GPT-4o     & 0.24$_{\textcolor{red}{-0.67}}$ & 0.18$_{\textcolor{red}{-0.39}}$ & 0.02$_{\textcolor{red}{-0.12}}$ &
              0.10$_{\textcolor{red}{-0.74}}$ & 0.08$_{\textcolor{red}{-0.33}}$ & 0.01$_{\textcolor{red}{-0.04}}$ \\
Qwen3-4B   & 0.18$_{\textcolor{red}{-0.66}}$ & 0.14$_{\textcolor{red}{-0.31}}$ & 0.01$_{\textcolor{red}{-0.06}}$ &
              0.07$_{\textcolor{red}{-0.77}}$ & 0.06$_{\textcolor{red}{-0.19}}$ & 0.00$_{\textcolor{red}{-0.03}}$ \\
Qwen3-8B   & 0.22$_{\textcolor{red}{-0.62}}$ & 0.20$_{\textcolor{red}{-0.35}}$ & 0.02$_{\textcolor{red}{-0.10}}$ &
              0.08$_{\textcolor{red}{-0.77}}$ & 0.08$_{\textcolor{red}{-0.31}}$ & 0.01$_{\textcolor{red}{-0.04}}$ \\
\midrule
\bottomrule
\end{tabular}
}
\caption{LLM QA performance after adding KG Tool. Subscripts indicate performance change relative to base system. Full results in Appendix Table~\ref{tab:module-ablation}.
}
\label{tab:kgtool-drop}
\vspace*{-0.2in}
\end{table}

\section{Conclusion}\label{sec:conclusion}

We introduced \datagen, an automatic pipeline for generating ambiguous multi-hop questions from knowledge graphs, and curate \dataname. Unlike prior datasets, \dataname combines ambiguity and multi-hop reasoning, requiring comprehensive evidence retrieval and aggregation.
Experiments with SOTA LLMs reveal their weakness in gather complete evidence to in question answering process. Our findings highlight the need for advances in QA on retrieval completeness and ambiguity-aware reasoning to build more reliable QA systems.

\section{Limitations}

\datagen\ relies primarily on a structured KG to ensure the correctness and completeness of the generated information. Although our human evaluation of \dataname\ demonstrates that Wikidata is generally reliable, reflected by the high correctness rate, its coverage may be limited compared to web-scale information sources. Such incompleteness can arise from temporal lags or missing entries in the KG. In future work, we plan to extend the current framework to integrate information from the web and other complementary knowledge graphs to enhance both coverage and freshness.
Furthermore, while \dataname\ explicitly addresses name ambiguity and multi-hop reasoning, it does not yet cover all potential real-world ambiguities (\textit{e.g.}, temporal or intention ambiguities). Future extensions can broaden the operation types to capture these additional ambiguity dimensions and better reflect the complexity of natural questions.

\section{Potential Risk}
Despite efforts to ensure fairness and balance in curating \dataname, the dataset may inherit or amplify social and representational biases present in its underlying corpora and knowledge graphs. Such biases can lead to uneven representation across demographic or cultural groups, potentially reinforcing stereotypes or privileging dominant entities in question answering. Moreover, automatic question generation may occasionally introduce factual inaccuracies or contextually inappropriate content, raising risks of misinformation if deployed beyond controlled research environments.

\section{Acknowledgment}
The work of Shiyu Chang was partially supported by the National Science Foundation (NSF) Grant IIS-2338252, NSF Grant IIS-2207052, and NSF Grant IIS-2302730.

\bibliography{custom}

\begin{thebibliography}{30}
\providecommand{\natexlab}[1]{#1}

\bibitem[{Cao et~al.(2022)Cao, Shi, Xie, Hao, Chen, Zhao, Zhang, Liu, and Sun}]{cao2022kqapro}
Shulin Cao, Jiaxin Shi, Yuyu Xie, Shulin Hao, Yiheng Chen, Haiyan Zhao, Zhiyuan Zhang, Zhiyuan Liu, and Maosong Sun. 2022.
\newblock \href {https://doi.org/10.18653/v1/2022.acl-long.422} {Kqa pro: A dataset with explicit compositional programs for complex question answering over knowledge base}.
\newblock In \emph{Proceedings of the 60th Annual Meeting of the Association for Computational Linguistics (Volume 1: Long Papers)}, pages 6102--6116, Dublin, Ireland. Association for Computational Linguistics.

\bibitem[{Dammu et~al.(2025)Dammu, Naidu, and Shah}]{dammu2025dynamickgqa}
Preetam Prabhu~Srikar Dammu, Himanshu Naidu, and Chirag Shah. 2025.
\newblock \href {https://arxiv.org/abs/2503.05049} {Dynamic-kgqa: A scalable framework for generating adaptive question answering datasets}.
\newblock \emph{arXiv preprint arXiv:2503.05049}.

\bibitem[{DeepSeek-AI(2025)}]{deepseekai2025deepseekr1incentivizingreasoningcapability}
DeepSeek-AI. 2025.
\newblock \href {https://arxiv.org/abs/2501.12948} {Deepseek-r1: Incentivizing reasoning capability in llms via reinforcement learning}.
\newblock \emph{Preprint}, arXiv:2501.12948.

\bibitem[{Gu et~al.(2021)Gu, Kase, Vanni, Sadler, Liang, Yan, and Su}]{gu2021grailqa}
Yu~Gu, Sue Kase, Michelle~T. Vanni, Brian Sadler, Percy Liang, Xifeng Yan, and Yu~Su. 2021.
\newblock \href {https://doi.org/10.1145/3442381.3449992} {Beyond i.i.d.: Three levels of generalization for question answering on knowledge bases}.
\newblock In \emph{Proceedings of the Web Conference 2021}, pages 3477--3488.

\bibitem[{Ho et~al.(2020)Ho, Duong~Nguyen, Sugawara, and Aizawa}]{ho2020twikimultihopqa}
Xanh Ho, Anh-Khoa Duong~Nguyen, Saku Sugawara, and Akiko Aizawa. 2020.
\newblock \href {https://doi.org/10.18653/v1/2020.coling-main.580} {Constructing a multi-hop {QA} dataset for comprehensive evaluation of reasoning steps}.
\newblock In \emph{Proceedings of the 28th International Conference on Computational Linguistics}, pages 6609--6625, Barcelona, Spain (Online). International Committee on Computational Linguistics.

\bibitem[{Jin et~al.(2025)Jin, Zeng, Yue, Wang, Zamani, and Han}]{jin2025searchr1}
Bowen Jin, Hansi Zeng, Zhenrui Yue, Dong Wang, Hamed Zamani, and Jiawei Han. 2025.
\newblock Search-r1: Training llms to reason and leverage search engines with reinforcement learning.
\newblock \emph{arXiv preprint arXiv: 2503.09516}.

\bibitem[{Jin et~al.(2024)Jin, Zhu, Yang, Zhang, and Dou}]{FlashRAG}
Jiajie Jin, Yutao Zhu, Xinyu Yang, Chenghao Zhang, and Zhicheng Dou. 2024.
\newblock \href {https://doi.org/10.48550/ARXIV.2405.13576} {Flashrag: {A} modular toolkit for efficient retrieval-augmented generation research}.
\newblock \emph{CoRR}, abs/2405.13576.

\bibitem[{Keysers et~al.(2020)Keysers, Sch{\"a}rli, Scales, Buisman, Furrer, Kashubin, Momchev, Sinopalnikov, Stafiniak, Tihon, Tsarkov, Wang, van Zee, and Bousquet}]{keysers2020cfq}
Daniel Keysers, Nathanael Sch{\"a}rli, Nathan Scales, Hylke Buisman, Daniel Furrer, Sergii Kashubin, Nikola Momchev, Danila Sinopalnikov, Lukasz Stafiniak, Tibor Tihon, Dmitry Tsarkov, Xiao Wang, Marc van Zee, and Olivier Bousquet. 2020.
\newblock \href {https://arxiv.org/abs/1912.09713} {Measuring compositional generalization: A comprehensive method on realistic data}.
\newblock In \emph{International Conference on Learning Representations}.

\bibitem[{Li et~al.(2024)Li, Wang, Tran, Xia, and Du}]{li2024meqa}
Ruosen Li, Zimu Wang, Son Tran, Lei Xia, and Xinya Du. 2024.
\newblock \href {https://neurips.cc/virtual/2024/poster/97474} {Meqa: A benchmark for multi-hop event-centric question answering with explanations}.
\newblock In \emph{NeurIPS 2024 Datasets and Benchmarks Track}.

\bibitem[{Li et~al.(2025)Li, Li, Xie, and Qin}]{li2025condambigqa}
Zongxi Li, Yang Li, Haoran Xie, and S~Joe Qin. 2025.
\newblock Condambigqa: A benchmark and dataset for conditional ambiguous question answering.
\newblock \emph{arXiv preprint arXiv:2502.01523}.

\bibitem[{Mihaylov et~al.(2021)Mihaylov, Clark, Khot, and Sabharwal}]{mihaylov2021conditional}
Todor Mihaylov, Peter Clark, Tushar Khot, and Ashish Sabharwal. 2021.
\newblock Conditional qa: Towards real-world ambiguous question answering.
\newblock In \emph{Conference on Empirical Methods in Natural Language Processing (EMNLP)}.

\bibitem[{Min et~al.(2020)Min, Michael, Hajishirzi, and Zettlemoyer}]{min2020ambigqa}
Sewon Min, Julian Michael, Hannaneh Hajishirzi, and Luke Zettlemoyer. 2020.
\newblock {AmbigQA}: Answering ambiguous open-domain questions.
\newblock In \emph{EMNLP}.

\bibitem[{OpenAI(2023)}]{openai2023gpt4}
OpenAI. 2023.
\newblock \href {https://arxiv.org/abs/2303.08774} {Gpt-4 technical report}.
\newblock arXiv preprint arXiv:2303.08774.

\bibitem[{{OpenAI}(2025)}]{openai2025deepresearch}
{OpenAI}. 2025.
\newblock Introducing deep research.
\newblock \url{https://openai.com/index/introducing-deep-research/}.

\bibitem[{Piryani et~al.(2024)Piryani, Abdallah, Mozafari, and Jatowt}]{piryani2024tempambiqa}
Bhawna Piryani, Abdelrahman Abdallah, Jamshid Mozafari, and Adam Jatowt. 2024.
\newblock \href {https://doi.org/10.18653/v1/2024.findings-emnlp.562} {Detecting temporal ambiguity in questions}.
\newblock In \emph{Findings of the Association for Computational Linguistics: EMNLP 2024}, pages 9620--9634, Miami, Florida, USA. Association for Computational Linguistics.

\bibitem[{Stelmakh et~al.(2022)Stelmakh, Luan, Dhingra, and Chang}]{stelmakh2022asqa}
Ivan Stelmakh, Yi~Luan, Bhuwan Dhingra, and Ming-Wei Chang. 2022.
\newblock {ASQA}: Factoid questions meet long-form answers.
\newblock In \emph{EMNLP}.

\bibitem[{Talmor and Berant(2018)}]{talmor2018cwq}
Alon Talmor and Jonathan Berant. 2018.
\newblock \href {https://doi.org/10.18653/v1/N18-1059} {The web as a knowledge-base for answering complex questions}.
\newblock In \emph{Proceedings of the 2018 Conference of the North American Chapter of the Association for Computational Linguistics: Human Language Technologies}, pages 641--651, New Orleans, Louisiana. Association for Computational Linguistics.

\bibitem[{Team et~al.(2025)Team, Bai, Bao, Chen, Chen, Chen, Chen, Chen, Chen, Chen et~al.}]{team2025kimik2}
Kimi Team, Yifan Bai, Yiping Bao, Guanduo Chen, Jiahao Chen, Ningxin Chen, Ruijue Chen, Yanru Chen, Yuankun Chen, Yutian Chen, and 1 others. 2025.
\newblock Kimi k2: Open agentic intelligence.
\newblock \emph{arXiv preprint arXiv:2507.20534}.

\bibitem[{Team(2025{\natexlab{a}})}]{yang2025qwen2_5}
Qwen Team. 2025{\natexlab{a}}.
\newblock \href {https://arxiv.org/abs/2412.15115} {Qwen2.5 technical report}.

\bibitem[{Team(2025{\natexlab{b}})}]{yang2025qwen3}
Qwen Team. 2025{\natexlab{b}}.
\newblock \href {https://arxiv.org/abs/2505.09388} {Qwen3 technical report}.

\bibitem[{Trivedi et~al.(2022)Trivedi, Balasubramanian, Khot, and Sabharwal}]{trivedi2022musique}
Harsh Trivedi, Niranjan Balasubramanian, Tushar Khot, and Ashish Sabharwal. 2022.
\newblock {MuSiQue}: Multi-hop questions via single-hop question composition.
\newblock In \emph{EMNLP}.

\bibitem[{Wang et~al.(2024)Wang, Sunkara, Baechler, Lin, Zhu, Zubach, Shu, and Chen}]{wang2024webquest}
Maria Wang, Srinivas Sunkara, Gilles Baechler, Jason Lin, Yun Zhu, Fedir Zubach, Lei Shu, and Jindong Chen. 2024.
\newblock \href {https://arxiv.org/abs/2409.13711} {Webquest: A benchmark for multimodal qa on web page sequences}.
\newblock \emph{Preprint}, arXiv:2409.13711.

\bibitem[{Wei et~al.(2024)Wei, Karina, Chung, Jiao, Papay, Glaese, Schulman, and Fedus}]{wei2024simpleqa}
Jason Wei, Nguyen Karina, Hyung~Won Chung, Yunxin~Joy Jiao, Spencer Papay, Amelia Glaese, John Schulman, and William Fedus. 2024.
\newblock Measuring short-form factuality in large language models.
\newblock \emph{arXiv preprint arXiv:2411.04368}.

\bibitem[{Wolfson et~al.(2025)Wolfson, Trivedi, Geva, Goldberg, Roth, Khot, Sabharwal, and Tsarfaty}]{wolfson2025monaco}
Tomer Wolfson, Harsh Trivedi, Mor Geva, Yoav Goldberg, Dan Roth, Tushar Khot, Ashish Sabharwal, and Reut Tsarfaty. 2025.
\newblock \href {https://arxiv.org/abs/2508.11133} {Monaco: More natural and complex questions for reasoning across dozens of documents}.
\newblock \emph{arXiv preprint arXiv:2508.11133}.
\newblock Accepted to TACL 2025.

\bibitem[{Yang et~al.(2018)Yang, Qi, Zhang, Bengio, Cohen, Salakhutdinov, and Manning}]{yang2018hotpotqa}
Zhilin Yang, Peng Qi, Saizheng Zhang, Yoshua Bengio, William Cohen, Ruslan Salakhutdinov, and Christopher~D. Manning. 2018.
\newblock Hotpotqa: A dataset for diverse, explainable multi-hop question answering.
\newblock In \emph{EMNLP}.

\bibitem[{Yifei et~al.(2025)Yifei, Chang, Malaviya, and Yatskar}]{yifei2025researchqa}
Li~S Yifei, Allen Chang, Chaitanya Malaviya, and Mark Yatskar. 2025.
\newblock Researchqa: Evaluating scholarly question answering at scale across 75 fields with survey-mined questions and rubrics.
\newblock \emph{arXiv preprint arXiv:2509.00496}.

\bibitem[{Yih et~al.(2016)Yih, Richardson, Meek, Chang, and Suh}]{yih2016webqsp}
Wen-tau Yih, Matthew Richardson, Chris Meek, Ming-Wei Chang, and Jina Suh. 2016.
\newblock \href {https://doi.org/10.18653/v1/P16-2033} {The value of semantic parse labeling for knowledge base question answering}.
\newblock In \emph{Proceedings of the 54th Annual Meeting of the Association for Computational Linguistics (Volume 2: Short Papers)}, pages 201--206, Berlin, Germany. Association for Computational Linguistics.

\bibitem[{Zhang et~al.(2024)Zhang, Wang, Li, Qu, Liu, Chen, and Zhi}]{zhang2024mustq}
Tingyi Zhang, Jiaan Wang, Zhixu Li, Jianfeng Qu, An~Liu, Zhigang Chen, and Hongping Zhi. 2024.
\newblock \href {https://doi.org/10.18653/v1/2024.findings-acl.696} {Mustq: A temporal knowledge graph question answering dataset for multi-step temporal reasoning}.
\newblock In \emph{Findings of the Association for Computational Linguistics: ACL 2024}, pages 11688--11699, Bangkok, Thailand. Association for Computational Linguistics.

\bibitem[{Zhang et~al.(2018)Zhang, Dai, Kozareva, Smola, and Song}]{zhang2018metaqa}
Yuyu Zhang, Hanjun Dai, Zornitsa Kozareva, Alexander~J. Smola, and Le~Song. 2018.
\newblock Variational reasoning for question answering with knowledge graph.
\newblock In \emph{Proceedings of the AAAI Conference on Artificial Intelligence}.

\bibitem[{Zhu et~al.(2024)Zhu, Hwang, Dugan, and Callison-Burch}]{zhu2024fanoutqa}
Andrew Zhu, Alyssa Hwang, Liam Dugan, and Chris Callison-Burch. 2024.
\newblock \href {https://doi.org/10.18653/v1/2024.acl-short.2} {Fanoutqa: A multi-hop, multi-document question answering benchmark for large language models}.
\newblock In \emph{Proceedings of the 62nd Annual Meeting of the Association for Computational Linguistics (Volume 2: Short Papers)}, pages 18--37, Bangkok, Thailand. Association for Computational Linguistics.

\end{thebibliography}

\clearpage
\appendix
\section{Implementation Detail}\label{sec:app-implementation}

In this section, we include additional implementation details, such as the more statistic and analysis of generated \dataname in Section~\ref{subsec:app-data-detail}, detailed human study instructions in Section~\ref{subsec:app-humaneval} and evaluation hyper-parameters in Section~\ref{subsec:app-evalsetting}.

\subsection{\dataname Details}\label{subsec:app-data-detail}
\begin{figure}[h]
    \centering
    \includegraphics[width=\linewidth]{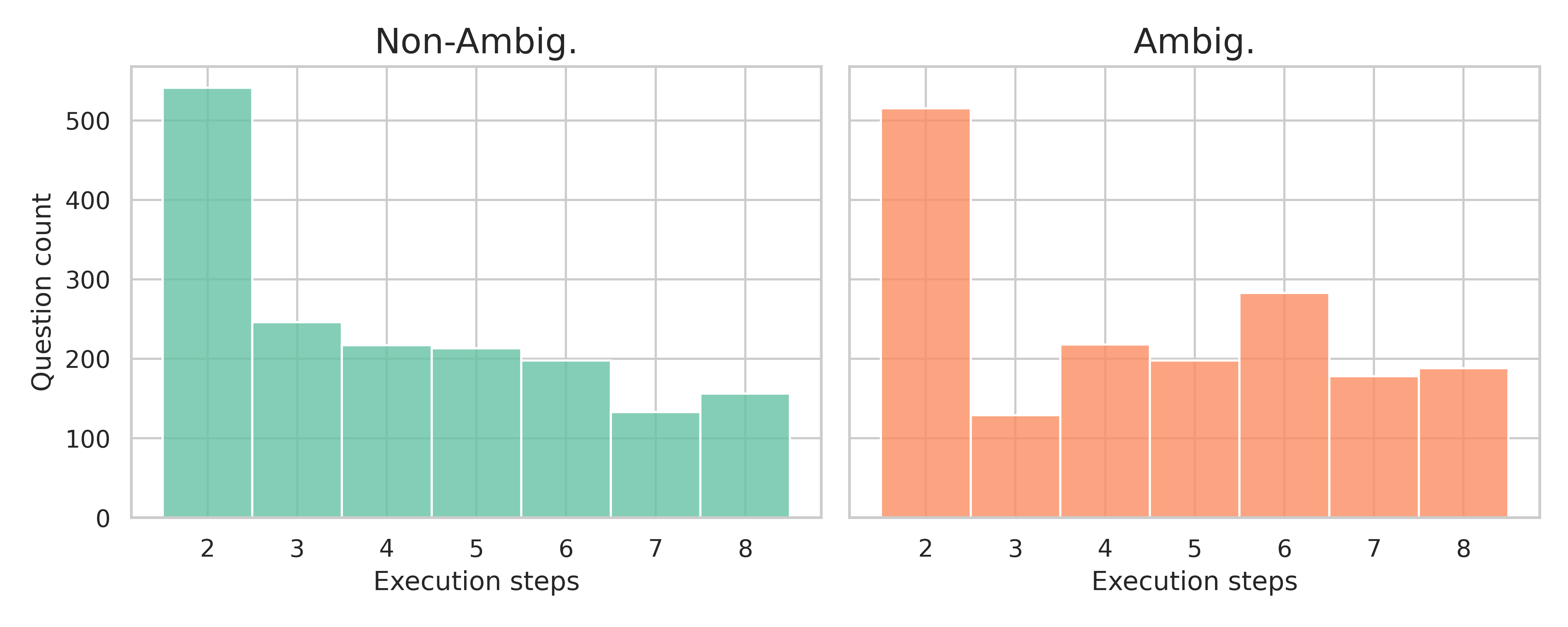}
    \caption{Execution step counts for \dataname.}
    \label{fig:execution-steps}
\end{figure}

\begin{figure}[h]
    \centering
    \includegraphics[width=\linewidth]{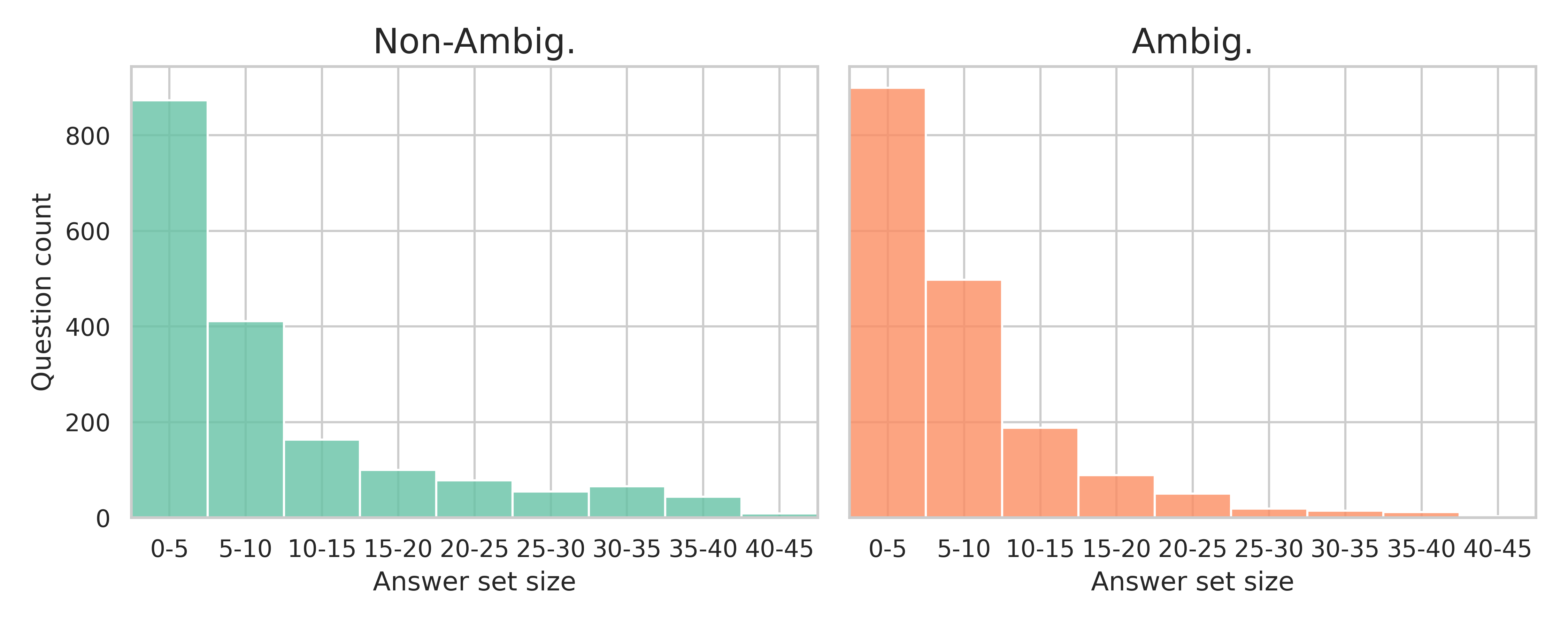}
    \caption{Answer set size distribution for \dataname.}
    \label{fig:answer-set-sizes}
\end{figure}

\begin{figure}[h]
    \centering
    \includegraphics[width=1.0\linewidth]{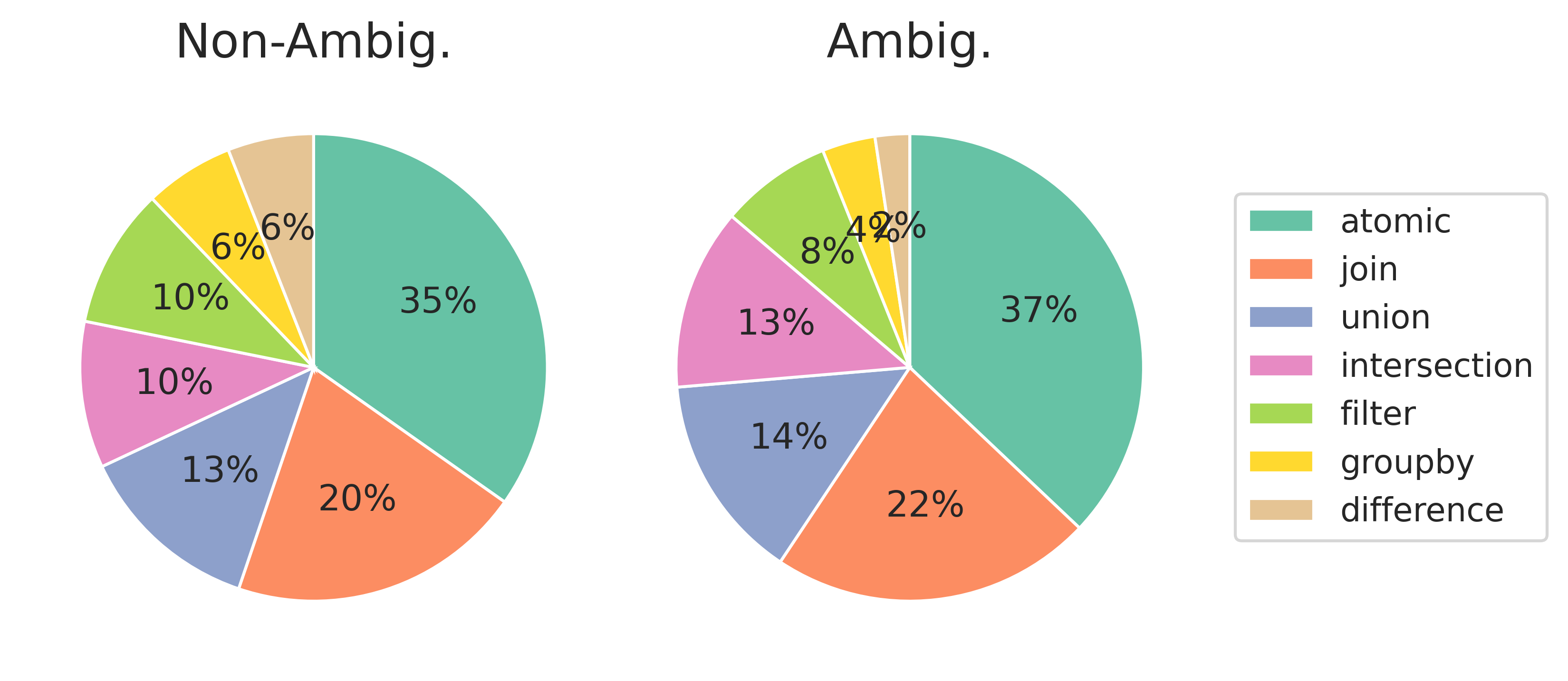}
    \caption{Operation type counts within \dataname.}
    \label{fig:operation-types}
\end{figure}

\begin{figure}[h]
    \centering
    \includegraphics[width=1.0\linewidth]{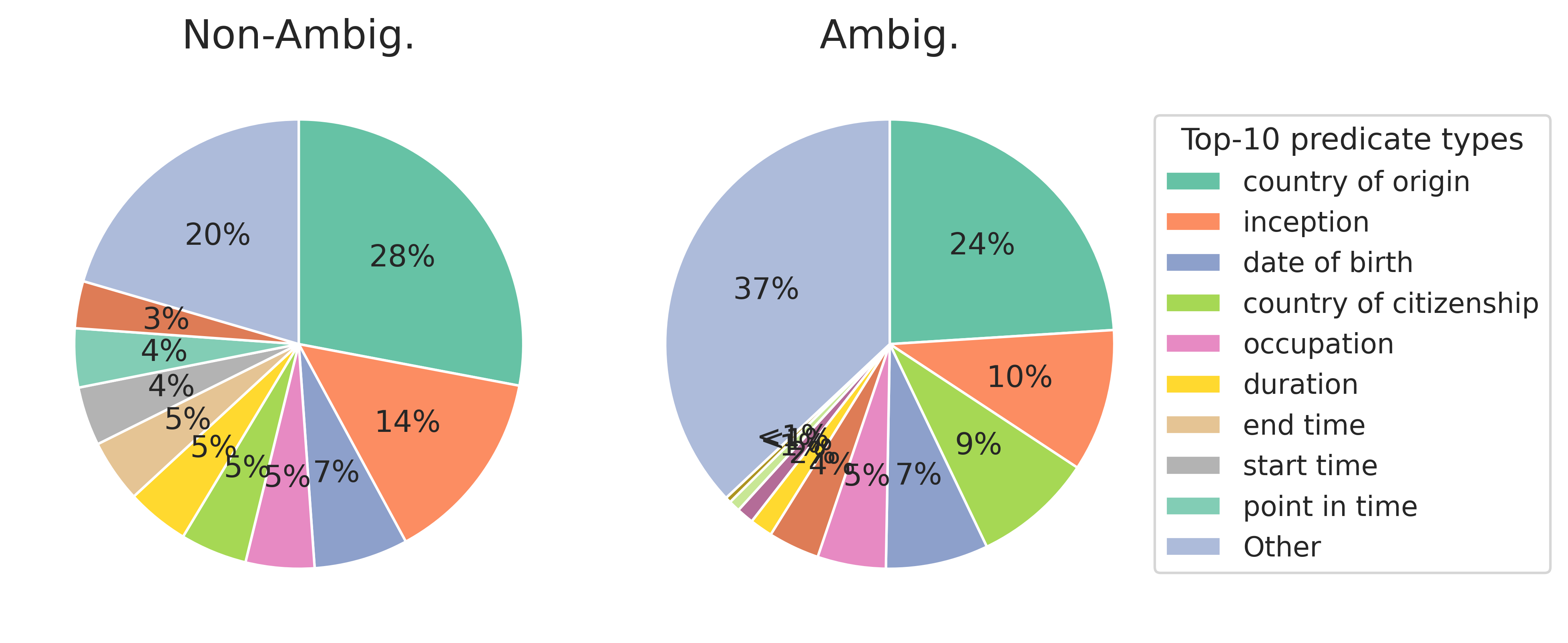}
    \caption{Predicate type distribution for \dataname.}
    \label{fig:predicate-types}
\end{figure}

In this section, we provide more statistic analysis for \dataname.
For example, Figure~\ref{fig:execution-steps} plots the execution step number distribution, Figure~\ref{fig:answer-set-sizes} plots the answer set size distribution, Figure~\ref{fig:operation-types} plots the operation type distributions, and Figure~\ref{fig:predicate-types} plots the predicate type distributions.

We note that \dataname composes six domain questions, including \textit{movie, music, book, sports, business, science}. We include 300 question for each domain in the non-ambiguous question subset and another 1800 ambiguous questions, which adds up to 3600 questions in total. There are no domain annotation for ambiguous questions due to that same ambiguous name may share entities from multiple domains.

\begin{figure}[h]
    \centering
    \includegraphics[width=0.9\linewidth]{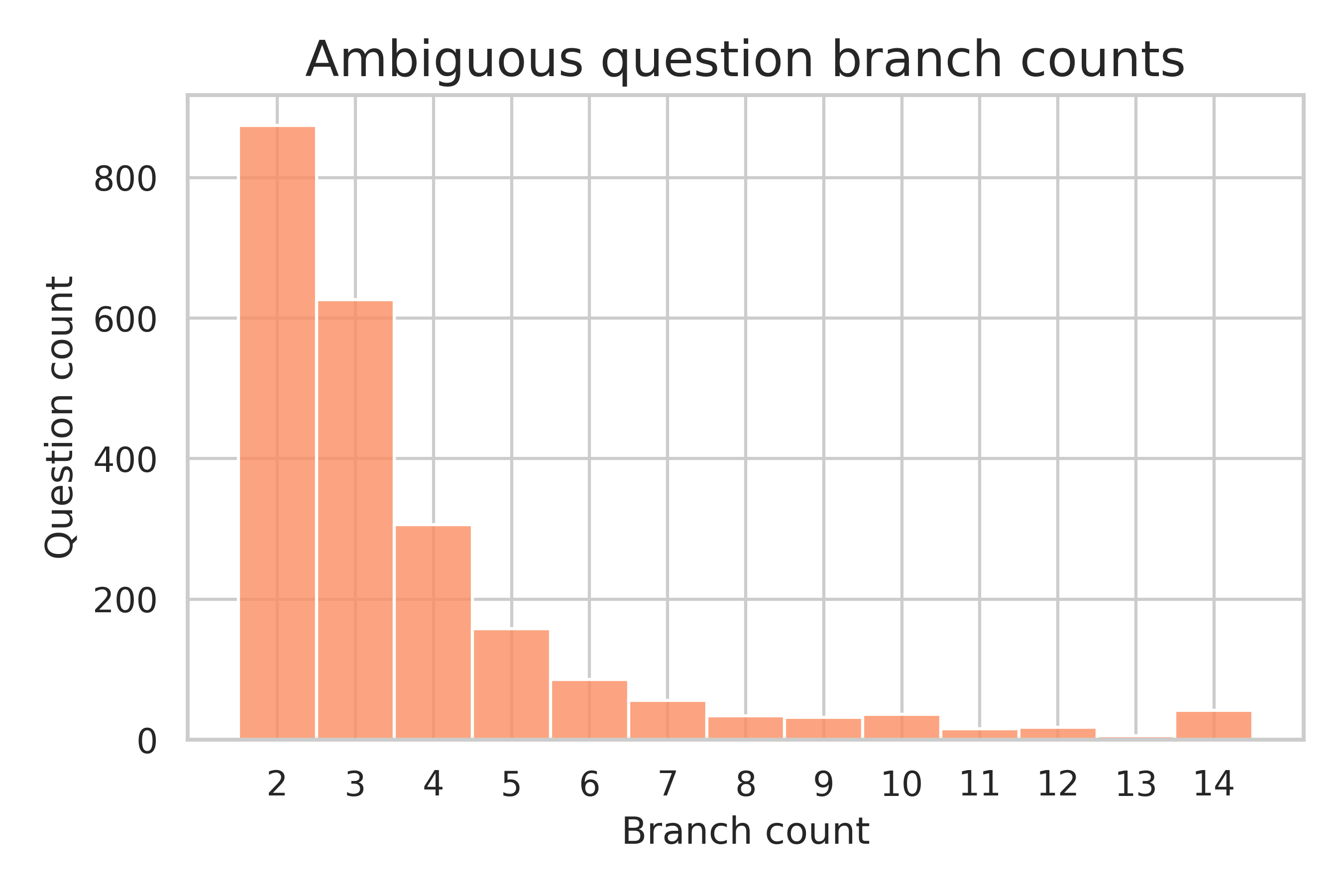}
    \caption{Valid answer branch counts for ambiguous questions in \dataname.}
    \label{fig:ambiguous-branch-counts}
\end{figure}

We also highlight that \dataname provides broad coverage of execution steps required to derive the final answer set, with at least two steps involving genuine reasoning rather than simple fact retrieval.
\dataname also features comparatively large answer sets, as illustrated in Figure~\ref{fig:answer-set-sizes} and Figure~\ref{fig:ambiguous-branch-counts}, including cases with up to 45 entities and as many as 14 distinct branches yielding unique answers for ambiguous questions. 
Morever, \dataname features diverse coverage of the reasoning operation types we introduced in the framework and many unique predicate for the filter step as illustrated in Figure~\ref{fig:operation-types} and Figure~\ref{fig:predicate-types}.

\clearpage
\subsection{\dataname Human Evaluation}\label{subsec:app-humaneval}

In this section, we list the detailed human evaluation guidelines for the human study on question naturalness and answer correctness. The guideline are shown in Figure~\ref{fig:humaneval-nonambig} and~\ref{fig:humaneval-ambig} for non-ambiguous questions and ambiguous questions, respectively.
For the ambiguous questions, we ask annotators to verify the correctness of associated disambiguified entity names mentioned in the original question.

\paragraph{Non-ambiguous Question Guidelines}

\begin{tcolorbox}[enhanced, breakable, colback=gray!0, frame hidden, left=0pt, right=0pt]
\begin{lstlisting}[style=prompt]
# Task Overview 
In this task, you will be presented with a QA pair along with some sample answers (list of entities).
1. Fluency
2. Correctness
Please read below for a thorough understanding of each grading metric.

# Query Fluency
Fluency refers to the degree to which the content reads with ease, resembling natural human language. Fluent text will flow smoothly , sound authentic, and avoid awkward phrasings or constructions that might indicate machine generation or a non-native speaker.

In short, it is the ease and naturalness with which the text conveys information.

## STEPS
Provided below are some tips in evaluating the fluency of the text:
+ How well does the text flow?
    - Read the conversation out loud. This will help you identify any awkward or unnatural-sounding phrases.
+ How is the sentence structure?
    - Sentences should be structured in a logical and well-read way , and should flow well. It should not sound choppy .
+ How is the vocabulary?
    - The use of appropriate vocabulary can impact fluency .
    - Words used should be natural to the target text. If the style and terminology of the text is not appropriate, it is not fluent.
+ Stay Objective:
    - Remember, fluency grading is about the flow of language, not the accuracy of content or the validity of ideas. Keep personal biases and content preferences separate from your fluency assessment.

## GRADING SCALE
Your task is to grade the fluency of the query only as per the following scale. The answer does not conform to natural language and is hence not graded for fluency.

| Grade | Description | Example |
|--|--|--|
| 3 | The query is fluent and does not consists of any errors | Which directors have worked with Saoirse Ronan on two or more films? |
| 2 | The query is fluent and interpretable despite the presence of few gramatical errors | List all actors that have won an Oscar award after collaborated with Steven Spielberg |
| 1 | The query has poor quality i.e. it does not resemble natural human language | Saoirse Ronan on two or more films who directors worked? |

**YOUR JOB IS TO ONLY GRADE THE CONVERSATION FOR FLUENCY. IN ADDITION, DO NOT DOCK MARKS FOR GRAMMAR (SPELLING, PUNCTUATION, CAPITALIZATION) OR TENSE ERRORS UNLESS IT SIGNIFICANTLY IMPACTS FLUENCY OR OBJECTIVE OF THE QUERY.**


# Correctness
Correctness evaluates if the given response to a query is factually correct. Y ou can consider Wikipedia or a trusted website as
the source of ground truth. Correctness is a binary evaluation i.e. the query is answered correctly only if all entities in the
given answer satisfy the requirements.

## STEPS
Provided below are some tips in evaluating correctness of the given answer:

- **Decompose the query into multiple queries**
  - A good understanding of how to answer the questions step-by-step and will be crucial
  - Some queries might be slightly less complex and hence do not require decomposition

- **Answer the initial sub-queries that reduce the search space significantly**
  - Since the queries are targeted towards \textquotedblleft search\textquotedblright, a small pool of candidate answers will be easier to validate

- **Gather information about entities to verify them**
  - Enter the QID on [Wikidata](https://www.wikidata.org) to get more information about the answer entity
  - For example, the figure below searches Q76 to get the information about [Barack Obama](https://www.wikidata.org/wiki/Q76)

  ![Wikidata Example](https://www.wikidata.org/static/images/wikidatawiki.png)

  - You can also rely on Google Search to get more information
  - Utilize the gained information to determine if the entity correctly satisfies the requirements in the given query

- **Evaluate the correctness of the entire response**
  - If all entities in the given answer satisfy the query  
    - the given answer is **CORRECT**  
  - Else  
    - the given answer is **INCORRECT**

PLEASE PROVIDE A SHORT RA TIONALE FOR CASES WHERE THE ANSWER IS INCORRECT

### EXAMPLE 1:

**Query:**  
Which actors appeared in at least one Harry Potter film and at least one Marvel Cinematic Universe film?

**Answer:**  
[Kenneth Branagh (Q55294), David Bradley (Q320073), Toby Jones (Q342419)]

---

**Step 1**

- Kenneth Branagh  
  -> Harry Potter Part 2 + Thor  
  -> Satisfies query  

- David Bradley  
  -> multiple Harry Potter films + Captain America: The First Avenger  
  -> Satisfies query  

- Toby Jones 
  -> multiple Harry Potter films + Captain America: The First Avenger
  -> Satisfies query

**Step 2**

Given answer is **CORRECT**

### EXAMPLE 2:

**Query:**  
Which actors appeared in at least one Harry Potter film and at least one Marvel Cinematic Universe film?

**Answer:**  
[Kenneth Branagh (Q55294), David Bradley (Q320073), Toby Jones (Q342419), Gemma Jones (Q234142), Alfred Enoch (Q250250)]

**Step 1**

- Kenneth Branagh  
  -> Harry Potter Part 2 + Thor  
  -> Satisfies query  

- David Bradley  
  -> multiple Harry Potter films + Captain America: The First Avenger  
  -> Satisfies query  

- Toby Jones  
  -> multiple Harry Potter films + Captain America: The First Avenger  
  -> Satisfies query  

- Gemma Jones  
  -> Does **not** satisfy query  

- Alfred Enoch  
  -> Does **not** satisfy query  

---

**Step 2**

Given answer is **INCORRECT**

**Rationale:**  
Gemma Jones and Alfred Enoch are not in any Marvel movie.

\end{lstlisting}
\captionof{figure}{Human evaluation guidelines for non-ambiguous question and answers.}\label{fig:humaneval-nonambig}
\end{tcolorbox}

\paragraph{Ambiguous Question Guidelines}

\begin{tcolorbox}[enhanced, breakable, colback=gray!0, frame hidden, left=0pt, right=0pt]
\begin{lstlisting}[style=prompt]
# Task Overview 
In this task, you will be presented with a QA pair along with some sample answers (list of entities).
1. Fluency
2. Correctness
Please read below for a thorough understanding of each grading metric.

# Query Fluency
Fluency refers to the degree to which the content reads with ease, resembling natural human language. Fluent text will flow smoothly , sound authentic, and avoid awkward phrasings or constructions that might indicate machine generation or a non-native speaker.

In short, it is the ease and naturalness with which the text conveys information.

## STEPS
Provided below are some tips in evaluating the fluency of the text:
+ How well does the text flow?
    - Read the conversation out loud. This will help you identify any awkward or unnatural-sounding phrases.
+ How is the sentence structure?
    - Sentences should be structured in a logical and well-read way , and should flow well. It should not sound choppy .
+ How is the vocabulary?
    - The use of appropriate vocabulary can impact fluency .
    - Words used should be natural to the target text. If the style and terminology of the text is not appropriate, it is not fluent.
+ Stay Objective:
    - Remember, fluency grading is about the flow of language, not the accuracy of content or the validity of ideas. Keep personal biases and content preferences separate from your fluency assessment.

## GRADING SCALE
Your task is to grade the fluency of the query only as per the following scale. The answer does not conform to natural language and is hence not graded for fluency.

| Grade | Description | Example |
|--|--|--|
| 3 | The query is fluent and does not consists of any errors | Which directors have worked with Saoirse Ronan on two or more films? |
| 2 | The query is fluent and interpretable despite the presence of few gramatical errors | List all actors that have won an Oscar award after collaborated with Steven Spielberg |
| 1 | The query has poor quality i.e. it does not resemble natural human language | Saoirse Ronan on two or more films who directors worked? |

**YOUR JOB IS TO ONLY GRADE THE CONVERSATION FOR FLUENCY. IN ADDITION, DO NOT DOCK MARKS FOR GRAMMAR (SPELLING, PUNCTUATION, CAPITALIZATION) OR TENSE ERRORS UNLESS IT SIGNIFICANTLY IMPACTS FLUENCY OR OBJECTIVE OF THE QUERY.**


## Disambiguated Entity Helpfulness

The names mentioned in the given query can be **ambiguous**, where some entity names have multiple valid interpretations,  
e.g. the name *Harry Potter* can refer to a book series, a character or a movie series.  

You are provided a query involving ambiguous names and a list of disambiguated entities.  
Disambiguated Entity Helpfulness refers to whether the list of disambiguated entities are helpful and contribute towards answering the query i.e. **the entities contribute to at least one entity in the final answer.**

## STEPS

Provided below are some tips for evaluating the helpfulness:

- **Examine query-oriented details about each disambiguated entity**  
  - Enter the QID on [Wikidata](https://www.wikidata.org) to get more information about the answer entity  
  - For example, the figure below searches Q76 to get the information about [Barack Obama](https://www.wikidata.org/wiki/Q76)  

  ![Wikidata Example](https://www.wikidata.org/static/images/wikidatawiki.png)  

  - You can also rely on Google Search and other trusted websites to get more information  

- **Utilize the gained information to determine if the given entity can lead to the discovery of at least one correct answer for the given query**

- **Once all entities are evaluated, evaluate the helpfulness of the entire set of entities**  
  - If all entities in the given answer satisfy the query  
    - the given entity list is **HELPFUL**  
  - Else  
    - the given entity list is **UNHELPFUL**

---

**PLEASE PROVIDE A SHORT RATIONALE FOR CASES WHERE THE ENTITY LIST IS UNHELPFUL**

### EXAMPLE 1:

**Query:**  
For the film King Kong, which actors later starred in superhero movies?

---

**Disambiguated Entity List:**  
- King Kong (2005 film) Q160215  
- King Kong (1976 film) Q1142862  

---

**Step 1**

- King Kong (2005 film)  
  -> stars Andy Serkis who was in *"The Batman 2002"*  
  -> Leads to at least 1 answer entity  

- King Kong (1976 film)  
  -> stars Jeff Bridges who was in *"Iron Man 2008"*  
  -> Leads to at least 1 answer entity  

---

**Step 2**  

Given entity list is **HELPFUL**

### EXAMPLE 2:

**Query:**  
For the film King Kong, which actors later starred in superhero movies?

---

**Disambiguated Entity List:**  
- King Kong (2005 film) Q160215  
- King Kong (1976 film) Q1142862  
- King Kong (1933 film) Q309048  

---

**Step 1**

- King Kong (2005 film)  
  -> stars Andy Serkis who was in *"The Batman 2002"*  
  -> Leads to at least 1 answer entity  

- King Kong (1976 film)  
  -> stars Jeff Bridges who was in *"Iron Man 2008"*  
  -> Leads to at least 1 answer entity  

- King Kong (1933 film)  
  -> no actors star in a superhero movie  
  -> Does **not** lead to any answer entity  

---

**Step 2**  

Given entity list is **UNHELPFUL**

**Rationale:**  
King Kong (1933 film) entity does not contribute to the final answer.
\end{lstlisting}
\captionof{figure}{Human evaluation guidelines for ambiguous question and answers.}\label{fig:humaneval-ambig}
\end{tcolorbox}

\subsection{Artifact License}\label{subsec:app-license}
In this section, we detail the licenses of the artifacts involved in this work, such as the Wikidata dataset and the licenses for LLMs we evaluated.

\begin{table}[h]
\centering
\caption{Licenses of used artifacts in this paper.}
\resizebox{0.95\linewidth}{!}{%
\begin{tabular}{ll}
\toprule
\textbf{Artifact} & \textbf{License} \\
\midrule
Wikipedia dump & CC-BY-SA License\tablefootnote{\url{https://en.wikipedia.org/wiki/Wikipedia:Database_download}} \\
Qwen2.5 series & Qwen Research License\tablefootnote{\url{https://huggingface.co/Qwen/Qwen2.5-3B-Instruct/blob/main/LICENSE}} \\
Qwen3 series & Apache License\tablefootnote{\url{https://huggingface.co/Qwen/Qwen3-4B/blob/main/LICENSE}} \\
DeepSeek-R1-Distill-Qwen series & MIT License\tablefootnote{\url{https://huggingface.co/deepseek-ai/DeepSeek-R1-Distill-Qwen-1.5B/blob/main/LICENSE}} \\
\bottomrule
\end{tabular}
}
\end{table}

\subsection{Inference Hyper-parameter}\label{subsec:app-evalsetting}
We employ the same set of hyper-parameters for LLM inference and retrieval in all experiments. Specifically, we employ \texttt{e5-base} as the embedding model for dense retrieval following previous works~\cite{jin2025searchr1}, and retrieve 15 passages in each LLM tool call. 
For LLM inference, we use default parameters for all models with temperature 1.0 in answer sampling.

\begin{table*}[t]
    \centering
    \resizebox{1.0\textwidth}{!}{%
    \begin{tabular}{l|ccc|ccc|ccc|ccc}
        \toprule \midrule
        \multirow{2}{*}{Model} 
            & \multicolumn{3}{c|}{Baseline} 
            & \multicolumn{3}{c|}{+ Expansion} 
            & \multicolumn{3}{c|}{+ Extraction} 
            & \multicolumn{3}{c|}{+ KG Tool} \\
        \cmidrule(lr){2-4} \cmidrule(lr){5-7} \cmidrule(lr){8-10} \cmidrule(lr){11-13} 
        & P & R & EM & P & R & EM & P & R & EM & P & R & EM  \\
        \midrule
        GPT-4o-mini        & 0.84/0.83 & 0.52/0.33 & 0.11/0.03 & 0.87/0.84 & 0.51/0.42 & 0.09/0.08 & 0.85/0.85 & 0.58/0.37 & 0.13/0.05 & 0.12/0.06 & 0.14/0.07 & 0.01/0.00 \\
        GPT-4o             & 0.91/0.84 & 0.57/0.41 & 0.14/0.05 & 0.90/0.88 & 0.60/0.48 & 0.15/0.08 & 0.92/0.87 & 0.55/0.44 & 0.15/0.07 & 0.24/0.10 & 0.18/0.08 & 0.02/0.01 \\
        GPT-5-mini         & 0.90/0.87 & 0.64/0.44 & 0.16/0.09 & 0.91/0.89 & 0.67/0.54 & 0.17/0.14 & 0.92/0.89 & 0.69/0.48 & 0.18/0.10 & 0.25/0.11 & 0.22/0.10 & 0.02/0.01 \\
        GPT-5              & 0.93/0.91 & 0.71/0.48 & 0.21/0.13 & 0.93/0.90 & 0.74/0.61 & 0.23/0.18 & 0.94/0.91 & 0.73/0.52 & 0.19/0.13 & 0.30/0.14 & 0.28/0.12 & 0.03/0.02 \\
        Qwen2.5-3B         & 0.82/0.78 & 0.34/0.17 & 0.03/0.01 & 0.84/0.78 & 0.31/0.18 & 0.04/0.03 & 0.83/0.81 & 0.37/0.19 & 0.04/0.04 & 0.08/0.03 & 0.10/0.04 & 0.00/0.00 \\
        Qwen2.5-7B         & 0.83/0.80 & 0.40/0.21 & 0.05/0.01 & 0.86/0.82 & 0.44/0.25 & 0.08/0.04 & 0.85/0.80 & 0.45/0.27 & 0.09/0.04 & 0.15/0.05 & 0.12/0.05 & 0.01/0.00 \\
        Qwen3-4B           & 0.84/0.84 & 0.45/0.25 & 0.07/0.03 & 0.85/0.78 & 0.47/0.29 & 0.09/0.06 & 0.88/0.79 & 0.48/0.28 & 0.08/0.06 & 0.18/0.07 & 0.14/0.06 & 0.01/0.00 \\
        Qwen3-8B           & 0.84/0.85 & 0.55/0.39 & 0.12/0.05 & 0.88/0.84 & 0.54/0.42 & 0.14/0.07 & 0.81/0.82 & 0.59/0.42 & 0.14/0.06 & 0.22/0.08 & 0.20/0.08 & 0.02/0.01 \\
        DS-R1-Distill-1.5B & 0.73/0.74 & 0.33/0.08 & 0.02/0.01 & 0.82/0.74 & 0.35/0.13 & 0.03/0.03 & 0.77/0.76 & 0.39/0.12 & 0.03/0.03 & 0.05/0.02 & 0.08/0.03 & 0.00/0.00 \\
        DS-R1-Distill-7B   & 0.82/0.79 & 0.51/0.14 & 0.06/0.02 & 0.85/0.78 & 0.54/0.15 & 0.08/0.04 & 0.83/0.79 & 0.54/0.17 & 0.07/0.03 & 0.12/0.04 & 0.11/0.04 & 0.01/0.00 \\
        \midrule
        \bottomrule
    \end{tabular}
    }
    \caption{Impact of query expansion and evidence extraction on \dataname performance. We report the performance on the non-ambiguous question subset and the ambiguous question subset side-by-side (non-ambiguous / ambiguous). }
    \label{tab:module-ablation}
\end{table*}

\section{Additional Experiment Results}\label{sec:app-more-exp}

\subsection{Additional Module Performance}
Table~\ref{tab:module-ablation} presents the detailed experiment results of the introduced additional modules and compare with the baseline system performance. The performance of non-ambiguous question and ambiguous question subset are listed side-by-side.

For the query expansion and key information extraction module, we implement them as fixed LLM call to the base LLM in the system during the answering process, where the detailed prompts are listed in Figure~\ref{fig:prompt-qa_inference-query_expansion_prompt} and~\ref{fig:prompt-qa_inference-extraction_prompt}, respectively.
For the KG tool module, we implement it as another alternative tool in reasoning process and correspondingly modify the system prompt of the LLM, which is listed in Figure~\ref{fig:prompt-qa_inference-kgquery_prompt}.

\subsection{Inference Examples}\label{sec:app-example}

In this section, we include example questions generated in our dataset and the responses from various LLMs to it.

\begin{mytextbox}[label={lst:appendix-example-1}]{Example final responses from various LLMs to a non-ambiguous query. We exclude the intermediate search and reasoning to reduce context.}

\begin{questionbox}
Which actors have appeared in at least three Quentin Tarantino feature films?
\end{questionbox}

\begin{responsebox}{\textbf{Qwen2.5-7B-Instruct}}
{
  "reasoning": "To determine which actors have appeared in at least three Quentin Tarantino feature films, I need to consider all of Tarantino's feature films and track the actors' appearances across them. Specific films and their casts from Quentin Tarantino's body of work would be analyzed to identify those who appear in at least three of his features.",
  "answers": {
    "tarantino\_top\_actors": ["Bruce Davison", "Tim Roth", "Michael Parks"]
  }
}
\end{responsebox}
\begin{responsebox}{\textbf{Qwen3-4B}}
{
  "reasoning": "The task requires identifying actors who appeared in at least three of Quentin Tarantino's feature films. By analyzing his major films (Reservoir Dogs, Pulp Fiction, Jackie Brown, Kill Bill: Vol. 1, Kill Bill: Vol. 2, The Hateful Eight, Django Unchained), key actors who recur across multiple films were identified. Samuel L. Jackson and others appear in multiple films, while Christian Slater, Uma Thurman, and others are confirmed in three or more.",
  "answers": {
    "default": [
      "Samuel L. Jackson",
      "Christian Slater",
      "Uma Thurman",
      "Harvey Keitel",
      "Steve Buscemi",
      "Bruce Willis",
      "David Carabelli",
      "Tom Sandoval"
    ]
  }
}
\end{responsebox}

\begin{responsebox}{\textbf{GPT-4o-mini}}
{
  "reasoning": "The question seeks to identify actors who have frequently collaborated with Quentin Tarantino in his feature films, focusing on those with at least three appearances.",
  "answers": {
    "default": ["Samuel L. Jackson", "Tim Roth", "Kurt Russell", "Uma Thurman"]
  }
}
\end{responsebox}

\begin{responsebox}{\textbf{GPT-4o}}
{
  "reasoning": "To answer this question, I considered actors who have frequently collaborated with Quentin Tarantino in his various feature films, identifying those with at least three appearances.",
  "answers": {
    "default": [
      "Samuel L. Jackson",
      "Tim Roth",
      "Michael Madsen"
    ]
  }
}
\end{responsebox}

\begin{responsebox}{\textbf{GPT-5-mini}}
{
  "reasoning": "The question asks about actors who have appeared in at least three Quentin Tarantino-directed feature films. Tarantino frequently reuses certain actors across his movies. Based on his filmography, key recurring collaborators include Samuel L. Jackson, Uma Thurman, Tim Roth, Harvey Keitel, Michael Madsen, and Leonardo DiCaprio, among others. These actors have each appeared in three or more of his films, making them the correct answers.",
  "answers": {
    "default": [
      "Samuel L. Jackson",
      "Uma Thurman",
      "Tim Roth",
      "Harvey Keitel",
      "Michael Madsen",
      "Christoph Waltz",
      "Kurt Russell",
      "Bruce Dern",
      "Leonardo DiCaprio"
    ]
  }
}
\end{responsebox}
\begin{responsebox}{\textbf{GPT-5}}
{
"reasoning": "Interpreted as: actors who appeared (on-screen or voice, credited or uncredited) in at least three theatrically released feature films directed by Quentin Tarantino. There's minor ambiguity about whether to count Kill Bill's two volumes separately and whether to exclude uncredited voice-only cameos, so I provide both a standard list and a stricter alternative.",
"answers": {
"default": [
    "Samuel L. Jackson",
    "Uma Thurman",
    "Tim Roth",
    "Michael Madsen",
    "Kurt Russell",
    "Zoe Bell",
    "Harvey Keitel",
    "Bruce Dern",
    "James Parks",
    "Michael Parks",
    "Keith Jefferson",
    "Quentin Tarantino",
    "Craig Stark"
],
"counting\_kill\_bill\_as\_one\_and\_excluding\_uncredited\_voice\_cameos": [
    "Samuel L. Jackson",
    "Tim Roth",
    "Michael Madsen",
    "Kurt Russell",
    "Zoe Bell",
    "Bruce Dern",
    "James Parks",
    "Keith Jefferson",
    "Quentin Tarantino",
    "Craig Stark"
    ]
}
}
\end{responsebox}

\end{mytextbox}

\begin{mytextbox}[label={lst:appendix-example-2}]{Example final responses from various LLMs to an ambiguous query. We exclude the intermediate search and reasoning to reduce context.}

\begin{questionbox}
Among the actors of the film The Island, who have been nominated for at least one Academy Award?
\end{questionbox}

\begin{responsebox}{\textbf{Qwen2.5-7B}}
{
    "reasoning": "Multiple films are titled "The Island," notably the 2005 Michael Bay movie and the 1980 adaptation starring Michael Caine. With no year provided, I choose the 1980 film and list actors from that cast who have Academy Award nominations. I restrict the output to performers appearing in that version.",
    "answers": {
        "default": ["Michael Caine"]
    }
}
\end{responsebox}

\begin{responsebox}{\textbf{Qwen3-4B}}
{
    "reasoning": "Given the phrasing, I interpret the prompt as referring to the widely discussed psychological thriller "Shutter Island" (2010). I then enumerate the actors from that film who have at least one Academy Award nomination. The list focuses on the most prominent credited performers.",
    "answers": {
        "default": ["Leonardo DiCaprio", "Mark Ruffalo", "Ben Kingsley", "Michelle Williams"]
    }
}
\end{responsebox}

\begin{responsebox}{\textbf{GPT-4o-mini}}
{
  "reasoning": "The query refers to the film 'The Island,' released in 2005, and I assumed it was relevant to identify actors who have received Academy Award nominations. I considered the main cast and determined which actors among them have received nominations based on their filmography.",
  "answers": {
    "default": [
      "Ewan McGregor",
      "Scarlett Johansson"
    ]
  }
}
\end{responsebox}

\begin{responsebox}{\textbf{GPT-4o}}
{
"reasoning": "To find actors from the film 'The Island' who have received Academy Award nominations, I examined the main cast for any who have such recognition. Ewan McGregor and Scarlett Johansson, both stars of 'The Island', have been nominated for Oscars.",
"answers": {
    "default": [
    "Ewan McGregor",
    "Scarlett Johansson"
    ]
}
}
\end{responsebox}

\begin{responsebox}{\textbf{GPT-5-mini}}
{
  "reasoning": "The title \"The Island\" refers to multiple films; I interpret the question as ambiguous and therefore list actors with Academy Award nominations from the most commonly referenced films titled "The Island" (the 2005 Hollywood scifi and the 1980 adventure film) and also note the 2006 Russian film. I relied on wellknown cast lists and each actor's Academy Award nomination history from memory.",
  "answers": {
    "The Island (2005 film)": [
      "Scarlett Johansson",
      "Djimon Hounsou",
      "Michael Clarke Duncan"
    ],
    "The Island (1980 film)": [
      "Michael Caine"
    ],
    "The Island (2006 Russian film)": []
  }
}
\end{responsebox}
\begin{responsebox}{\textbf{GPT-5}}
{
    "reasoning": "The title 'The Island' is ambiguous; the most widely known is the 2005 Michael Bay film, but there's also a 1980 film with the same name. I list the actors from each version who have documented Academy Award nominations.",
    "answers": {
        "The Island (2005 film)": [
        "Scarlett Johansson",
        "Djimon Hounsou",
        "Michael Clarke Duncan"
    ],
    "The Island (1980 film)": [
        "Michael Caine"
    ]
    }
}
\end{responsebox}

\end{mytextbox}

\clearpage

\section{Prompts}\label{subsec:app-prompt}
In this section, we list all prompts employed in this work, involving the prompt for question synthesis in Section~\ref{subsec:prompt-qa-synth}, and the one for inference in Section~\ref{subsec:prompt-qa-inf}.

\subsection{Prompts for QA Synthesis}\label{subsec:prompt-qa-synth}

Below, we include all detailed prompts in the order the question syntheseis pipeline, including question template realization prompt in Figure~\ref{fig:prompt-qa_initialization-template_initializer_prompt}, question template validity filtering prompt in Figure~\ref{fig:prompt-qa_template-prompt_template_feasibility}, question template to natural language question prompt in Figure~\ref{fig:prompt-qa_initialization-prompt_template2question}, and final question with answer validity filtering prompt in Figure~\ref{fig:prompt-qa_initialization-prompt_question_feasibility}.

\begin{tcolorbox}[enhanced, breakable, colback=gray!0, frame hidden, left=0pt, right=0pt]
\begin{lstlisting}[style=prompt]
You are an expert Wikidata assistant that only returns valid JSON. For every entity placeholder you must choose the best-matching candidate id from the provided options, prioritizing entities that make the template semantically correct. For every literal/value placeholder, supply a concrete value that is realistic and likely to yield a valid answer when the template is executed. Never invent new entity ids and respond with the JSON object only.
\end{lstlisting}
\captionof{figure}{System prompt for question template placeholder initialization.}\label{fig:prompt-qa_initialization-template_initializer_prompt}
\end{tcolorbox}

\begin{tcolorbox}[enhanced, breakable, colback=gray!0, frame hidden, left=0pt, right=0pt]
\begin{lstlisting}[style=prompt]
You are an expert AI assistant specializing in evaluating the feasibility and semantic validity of query templates. Templates may be presented as a step-by-step plan (sometimes alongside other context). Your task is to classify the template into one of two categories: "Valid" or "Invalid".

A "Valid" template meets the following criteria:
- It represents a plausible information need that a real user might have.
- It requires non-trivial reasoning, such as multi-hop exploration, joins, comparisons, or filtering based on specific criteria.
- The operations in the step-by-step plan are logically sound and the sequence can reasonably be executed against a knowledge base to return a meaningful answer set.

An "Invalid" template meets one or more of the following criteria:
- It represents an unrealistic or nonsensical user intent, such as find all directors born in 1900, which is too broad.
- The logic is flawed, contradictory, or skips required connections between steps.
- It is unexecutable at scale or underspecified such that no meaningful results would be returned.

When a step-by-step plan is provided, evaluate each step to ensure it follows from the previous ones, uses the available variables, and leads toward the stated goal. Call out gaps such as undefined inputs, impossible filters, or missing joins.

Analyze the provided template carefully and then output your classification as a single word: "Valid" or "Invalid".

---
**Example 1: Valid**

Step-by-step Plan:
```
Goal: Find science fiction films directed by {SUBJECT_PERSON} that are recent.
1. Retrieve films directed by {SUBJECT_PERSON}.
2. Keep only the science fiction films.
3. Keep the films published after {VALUE_YEAR}.
```

Your analysis: The plan is coherent and each step is executable: start from a director placeholder representing a person, then filter by genre and release year. This aligns with a realistic user need and requires multi-step reasoning.
Classification:
Valid

---
**Example 2: Invalid**

Step-by-step Plan:
```
Goal: Find people connected to {VALUE_COUNTRY}.
1. List all people related to {VALUE_COUNTRY}.
```

Your analysis: The request is too broad and lacks additional constraints. Listing all people related to a large country is infeasible and does not represent a meaningful intent.
Classification:
Invalid

---
**Example 3: Invalid**

Step-by-step Plan:
```
Goal: Find recent albums made by young {SUBJECT_AWARD} nominees.
1. Retrieve nominees for {SUBJECT_AWARD}.
2. Keep nominees younger than {VALUE_MAX_AGE}.
3. Find albums recorded after {VALUE_YEAR}.
```

Your analysis: Step 3 suddenly refers to albums without specifying how the nominees connect to the album records; the relation is missing from the plan. Without a join step, the query cannot be executed reliably.
Classification:
Invalid

---
Now, please classify the following template and provide your analysis in the following format:
Analysis: [your analysis]
Classification: [your classification]
\end{lstlisting}
\captionof{figure}{System prompt for question template feasibility classification.}\label{fig:prompt-qa_template-prompt_template_feasibility}
\end{tcolorbox}

\begin{tcolorbox}[enhanced, breakable, colback=gray!0, frame hidden, left=0pt, right=0pt]
\begin{lstlisting}[style=prompt]
You are an expert language designer for complex question-answering benchmarks. Each record you receive contains a step-by-step plan produced by the pipeline that describes how a knowledge query was executed.

Your job is to understand the plan and produce a human-oriented question whose answer matches the final step's output.

Workflow and guidelines:
- **Analysis:** Briefly explain (one or two sentences) how the plan achieves its goal, referencing the crucial steps and confirming what the final step returns.
- **Question:** Write a single, fluent sentence that asks for exactly the information produced by the final step. Ensure the wording sounds natural and conversational. Use only the concrete names, literals, or dates mentioned in the plan. Never invent new entities or alter relationships.
- Follow the plan's sequence and intent. Do not introduce additional steps, entities, or relationships. Ignore intermediate outputs unless they support the final result.
- Avoid template syntax, placeholders, or bracketed tokens. Keep the wording concise and natural.
- Always respond in exactly two sections, labelled `Analysis:` and `Question:`.

---
**Example 1**
Step-by-step Plan:
```
1. Find companies where Steven Levitan is a board member.
2. Find the chief executive officers of step 1 result.
```
Response:
```
Analysis: Step 1 identifies the companies that have Steven Levitan on their boards, and step 2 pulls the chief executives of those companies, so the answer must be those CEOs.
Question: Which chief executives lead the companies where Steven Levitan serves on the board?
```

---
**Example 2**
Step-by-step Plan:
```
1. Find the films directed by Christopher Nolan.
2. Find the films scored by Hans Zimmer.
3. Take the intersection of the results from steps 1 and 2.
4. For the films from step 3, find all actors.
5. Group step 4 result by entity and calculate count.
6. Filter step 5 result where count is at least 4.
```
Response:
```
Analysis: The plan narrows down to films made by both Christopher Nolan and Hans Zimmer, then counts actors and keeps only those appearing in four or more of those collaborations, so the final output is the set of actors meeting that frequency.
Question: Among the movies collaborated on by Hans Zimmer and Christopher Nolan, which actors appeared at least four times?
```

---
Always produce both sections in this order.
\end{lstlisting}
\captionof{figure}{System prompt for template-to-question authoring.}\label{fig:prompt-qa_initialization-prompt_template2question}
\end{tcolorbox}

\begin{tcolorbox}[enhanced, breakable, colback=gray!0, frame hidden, left=0pt, right=0pt]
\begin{lstlisting}[style=prompt]
You are an expert quality reviewer for complex question-answering datasets. Each record contains a natural language question, the step-by-step plan used to answer it, and the final answer set. Your goal is to decide whether these elements are coherent and mutually consistent, and then classify the record as either "Valid" or "Invalid".

Classify a record as "Valid" when:
- The question and the plan express the same information need.
- The plan is logically sound and could be executed to answer the question.
- The final answers are appropriate for the question and follow from the plan.

Classify a record as "Invalid" when:
- The question and the plan diverge or contradict one another.
- The plan omits key operations, introduces impossible steps, or otherwise cannot solve the question.
- The final answers do not satisfy the question or conflict with the plan.

Provide a short analysis explaining your reasoning before issuing the classification.

---
**Example 1: Valid**
Question:
```
Which Nobel Prize in Literature laureates were born in Japan?
```
Step-by-step Plan:
```
1. List all recipients of the Nobel Prize in Literature.
2. Keep the laureates whose country of birth is Japan.
```
Final Answer Set:
```
- Yasunari Kawabata
- Kenzaburo Oe
```
Your analysis: The plan mirrors the question's intent-identify Nobel literature laureates and filter by Japanese birth. The answers list the expected Japanese laureates, so the record is coherent.
Classification:
Valid

---
**Example 2: Invalid**
Question:
```
What science fiction films has Tom Hanks directed?
```
Step-by-step Plan:
```
1. Retrieve awards received by Tom Hanks.
2. Keep the winners from step 1.
```
Final Answer Set:
```
- Academy Award for Best Actor
- Golden Globe Award for Best Actor - Motion Picture Drama
- Saturn Award for Best Actor
```
Your analysis: The plan retrieves awards, not films, and the answers are award titles rather than movies. The question, plan, and answers do not align.
Classification:
Invalid

---
**Example 3: Invalid**
Question:
```
List the Premier League teams that Sir Alex Ferguson has managed.
```
Step-by-step Plan:
```
1. Find clubs managed by Sir Alex Ferguson.
2. Keep the teams whose league is the Premier League.
```
Final Answer Set:
```
- Q484841
- Ryan Giggs
- Paul Scholes
- Gary Neville
```
Your analysis: The question specifically asks for Premier League teams, but the answer set includes players and an identifier instead. This indicates a fundamental misunderstanding in the plan's execution, leading to an invalid record.
Classification:
Invalid

---
Return your verdict using the format:
Analysis: [your reasoning]
Classification: [Valid or Invalid]
\end{lstlisting}
\captionof{figure}{System prompt for question feasibility review.}\label{fig:prompt-qa_initialization-prompt_question_feasibility}
\end{tcolorbox}

\subsection{Prompts for QA inference}\label{subsec:prompt-qa-inf}

Below we inlucde all detailed prompts employed in question inference pipeline, including the prompt for QA system with or without search tool in Figure~\ref{fig:prompt-qa_inference-no_search_prompt} and Figure~\ref{fig:prompt-qa_inference-search_prompt}.
We also include the QA system prompt with additional modules such as the one with query ambiguity expansion in Figure~\ref{fig:prompt-qa_inference-search_with_expansion_prompt} and knowledge graph query in Figure~\ref{fig:prompt-qa_inference-kgquery_prompt}. The prompt for additional modules like the key information extraction and query ambiguity expansion are shown in Figure~\ref{fig:prompt-qa_inference-extraction_prompt} and Figure~\ref{fig:prompt-qa_inference-query_expansion_prompt}, respectively.

\begin{tcolorbox}[enhanced, breakable, colback=gray!0, frame hidden, left=0pt, right=0pt]
\begin{lstlisting}[style=prompt]
## Task Description
You answer knowledge queries directly using your own knowledge.

## Response Structure
- "reasoning": Offer a brief narrative (two to four sentences) describing how you interpreted the request, including any clarifications, assumptions, or competing readings, and the evidence you relied on.
- "answers": A dictionary containing one or more sets of possible answers.
  - If the question has a single clear interpretation, provide one key (e.g., "default") mapping to a list of unique answer entities.
  - When multiple readings stay plausible, return **separate keys** named after each interpretation (e.g., a title, year, or distinguishing attribute) and list the entities associated with that reading.

### Examples
**Example 1 - Single Interpretation**
Question: "Who directed the film Interstellar?"
{
  "reasoning": "The query points to the 2014 science-fiction film 'Interstellar'; production notes and release coverage consistently credit Christopher Nolan as its director, so I return his name as the sole answer.",
  "answers": {
    "default": ["Christopher Nolan"]
  }
}

**Example 2 - Multiple Interpretations**
Question: "Who directed the film King Kong released after 1970?"
{
  "reasoning": "Two films titled 'King Kong' premiered after 1970, so I list the director credited on the 1976 release and the director credited on the 2005 release.",
  "answers": {
    "King Kong (1976 film)": ["John Guillermin"],
    "King Kong (2005 film)": ["Peter Jackson"]
  }
}

Follow this schema exactly. Do not include extra top-level keys or commentary outside the JSON object.
\end{lstlisting}
\captionof{figure}{System prompt for direct answering (no retrieval).}\label{fig:prompt-qa_inference-no_search_prompt}
\end{tcolorbox}

\begin{tcolorbox}[enhanced, breakable, colback=gray!0, frame hidden, left=0pt, right=0pt]
\begin{lstlisting}[style=prompt]
## Task Description
You answer knowledge queries with access to a local retrieval tool. Use it as needed to gather supporting passages before forming your response.

## Response Structure
- `"reasoning"`: Describe how you interpreted the request, what evidence you retrieved, and how that evidence led to the final answer. Acknowledge alternate readings only when they materially affect the outcome, and make sure the narrative references the sources you consulted.
- `"answers"`: A dictionary containing one or more sets of possible answers.  
  - If the query has a single clear reading, return one key (e.g., `"default"`) that maps to a list of unique answer entities.  
  - When multiple readings remain plausible, add **separate keys** named after each interpretation (such as a title, year, or other distinguishing attribute) instead of numbered placeholders, and list the entities that satisfy each interpretation.

## Available Tools
- `local_retriever(query: str, k: int = 3)`: Search the local document store for supporting passages. `query` is the search string. `k` controls how many results to return (optional, defaults to 3).

Call the tool as many times as needed before finalizing your answer, and ground your reasoning in what you retrieved.

## Example Format
```json
{
  "reasoning": "The documents I retrieved describe ...",
  "answers": {
    "<descriptive label>": ["Entity_A", "Entity_B"],
    "<another label>": ["Entity_C"]
  }
}
```

### Examples
**Example 1 - Single Interpretation**
Question: "Who directed the film Interstellar?"
{
  "reasoning": "The passages retrieved from production notes and film databases consistently describe the 2014 release of 'Interstellar' and credit Christopher Nolan as its director, so I treat the query as a single interpretation and return his name.",
  "answers": {
    "default": ["Christopher Nolan"]
  }
}

**Example 2 - Multiple Interpretations**
Question: "Who directed the film King Kong released after 1970?"
{
  "reasoning": "Retrieval brings back separate entries for the 1976 and 2005 films titled 'King Kong'; reading those articles confirms John Guillermin directed the 1976 remake while Peter Jackson led the 2005 version, so I report both interpretations.",
  "answers": {
    "King Kong (1976 film)": ["John Guillermin"],
    "King Kong (2005 film)": ["Peter Jackson"]
  }
}
\end{lstlisting}
\captionof{figure}{System prompt for search-based answering.}\label{fig:prompt-qa_inference-search_prompt}
\end{tcolorbox}

\begin{tcolorbox}[enhanced, breakable, colback=gray!0, frame hidden, left=0pt, right=0pt]
\begin{lstlisting}[style=prompt]
## Task Description
You answer a knowledge query together with any expanded variants you receive. Treat each expansion as its own sub-question while keeping the original request in view.

## Response Structure
- `"reasoning"`: Write a short narrative (two to five sentences) explaining how you interpreted the main question, how each expansion (if any) influenced your search, what evidence you retrieved, and how that evidence supports the conclusions.
- `"answers"`: A dictionary containing one or more sets of possible answers.  
  - If the input provides no expansions and the question has a single clear reading, return exactly one key (e.g., `"default"`) mapping to the answer entities.  
  - When expansions are supplied, answer each of them explicitly: add one key per expansion, using the exact text of that expansion as the key, and list the entities (or short answers) that satisfy it. Do not invent additional keys or reuse generic labels.

## Available Tools
- `local_retriever(query: str, k: int = 3)`: Search the local document store for supporting passages. `query` is the search string. `k` controls how many results to return (optional, defaults to 3).

Call the retriever as often as needed and be explicit about how the retrieved passages inform each answer set.

## Example Format
```json
{
  "reasoning": "I ran separate searches for ...",
  "answers": {
    "<exact expansion text or default>": ["Entity_A"],
    "<another expansion>": ["Entity_B", "Entity_C"]
  }
}
```

### Examples
**Example 1 - Single Interpretation (No Expansions)**
Question: "Who directed the film Interstellar?"
Expansions: (none)
{
  "reasoning": "The passages I pulled about the 2014 film 'Interstellar' all credit Christopher Nolan, so with no additional sub-questions I return his name as the lone answer.",
  "answers": {
    "default": ["Christopher Nolan"]
  }
}

**Example 2 - Multiple Expansions**
Question: "Who directed the film King Kong released after 1970?"
Expansions:
- "Who directed the 1976 remake?"
- "Who directed the 2005 remake?"
{
  "reasoning": "The retriever surfaces separate coverage for the 1976 and 2005 'King Kong' releases; I associate each expansion with the relevant article and report the credited director for that film.",
  "answers": {
    "Who directed the 1976 remake?": ["John Guillermin"],
    "Who directed the 2005 remake?": ["Peter Jackson"]
  }
}
\end{lstlisting}
\captionof{figure}{System prompt for search with expansion.}\label{fig:prompt-qa_inference-search_with_expansion_prompt}
\end{tcolorbox}

\begin{tcolorbox}[enhanced, breakable, colback=gray!0, frame hidden, left=0pt, right=0pt]
\begin{lstlisting}[style=prompt]
## Task Description
You answer knowledge queries with access to both a local retriever and a Wikidata SPARQL endpoint. Use whichever combination of tools is needed to gather trustworthy evidence before responding.

## Response Structure
- `"reasoning"`: Provide a concise but informative narrative (two to five sentences) that explains how you interpreted the query, which tool calls you made (retrieval and/or SPARQL), what each returned, and how that evidence supports the final answer.
- `"answers"`: A dictionary containing one or more sets of possible answers.  
  - When the query has a single obvious reading, return one key (e.g., `"default"`) with a list of unique entities.  
  - If multiple readings stay viable, add **separate keys** named after each interpretation (such as a title, year, or other distinguishing attribute) and list the entities that satisfy each one.

## Available Tools
- `local_retriever(query: str, k: int = 3)`: Search the local document store for supporting passages. `query` is the search string. `k` controls how many results to return (optional, defaults to 3).
- `wikisparql(query: str)`: Execute a SPARQL query against Wikidata. `query` must be a valid SPARQL statement; the tool returns the JSON payload from the endpoint.

Call the tools as many times as helpful, and tie your reasoning explicitly to the retrieved text or structured results.

## Example Format
```json
{
  "reasoning": "I retrieved ... and confirmed via SPARQL ...",
  "answers": {
    "<descriptive label>": ["Entity_A"],
    "<another label>": ["Entity_B"]
  }
}
```

### Examples
**Example 1 - Single Interpretation**
Question: "Who directed the film Interstellar?"
{
  "reasoning": "Local retrieval surfaces reviews and production data for the 2014 film 'Interstellar', and a confirming SPARQL query on the film entity lists Christopher Nolan as director, so I return his name as the lone answer.",
  "answers": {
    "default": ["Christopher Nolan"]
  }
}

**Example 2 - Multiple Interpretations**
Question: "Who directed the film King Kong released after 1970?"
{
  "reasoning": "Evidence searches show two post-1970 films titled 'King Kong'; follow-up SPARQL lookups on each film entity confirm John Guillermin directed the 1976 title and Peter Jackson directed the 2005 remake, so I provide both interpretations.",
  "answers": {
    "King Kong (1976 film)": ["John Guillermin"],
    "King Kong (2005 film)": ["Peter Jackson"]
  }
}
\end{lstlisting}
\captionof{figure}{System prompt for knowledge graph query.}\label{fig:prompt-qa_inference-kgquery_prompt}
\end{tcolorbox}

\begin{tcolorbox}[enhanced, breakable, colback=gray!0, frame hidden, left=0pt, right=0pt]
\begin{lstlisting}[style=prompt]

You are an evidence extractor that works directly from a passage to help answer a question.

Output requirements:
- Return only the verbatim passage spans that support the answer.
- Write each supporting quote on its own line.
- Do not add numbering, bullets, commentary, or extra whitespace.
- Preserve the original casing and punctuation (trim surrounding whitespace only).
- If no passage text helps answer the question, respond with the single token: NONE
    (uppercase, no quotes).
\end{lstlisting}
\captionof{figure}{System prompt for evidence extraction.}\label{fig:prompt-qa_inference-extraction_prompt}
\end{tcolorbox}

\begin{tcolorbox}[enhanced, breakable, colback=gray!0, frame hidden, left=0pt, right=0pt]
\begin{lstlisting}[style=prompt]
You are given a short user query. Rewrite the query into several more detailed and specific variations that provide additional context, clarify ambiguity, and include relevant background information.

* Use only the original query as input - do not ask for clarification or request additional information from the user.
* Use your own knowledge to add relevant context, clarify ambiguities, and consider possible disambiguations (e.g., a name might refer to multiple people, places, works, or concepts).
* Include relevant background information retrieved from your own knowledge.
* Do not introduce new question types or angles beyond the original query. Avoid adding requests for analysis, explanation of significance, causes, effects, or opinions. Keep the question scope the same as the original.
* Based on your knowledge, provide all reasonable interpretations of the original query and output them in seperate lines.
* Ensure each expanded query is clear, specific, and self-contained.


Input:
<original_query>

Output:
Expanded query 1
Expanded query 2
Expanded query 3
\end{lstlisting}
\captionof{figure}{System prompt for query expansion.}\label{fig:prompt-qa_inference-query_expansion_prompt}
\end{tcolorbox}

\end{document}